\pdfoutput=1
\documentclass[11pt]{article}

\usepackage[]{acl}

\usepackage{times}
\usepackage{latexsym}

\usepackage[T1]{fontenc}
\usepackage[utf8]{inputenc}

\usepackage{microtype}

\usepackage{graphicx}
\usepackage{float}
\usepackage{todonotes} 
\usepackage{linguex}
\usepackage{soul}
\usepackage{booktabs}
\usepackage{caption}
\usepackage{subcaption}
\usepackage[inline]{enumitem}

\newcommand{\token}[1]{\textsc{#1}}
\def\newterm#1{\emph{#1}}

\title{Reconstruction Probing}

\author{Najoung Kim$,^{\dagger}$ Jatin Khilnani,$^{\Delta}$ Alex Warstadt,$^{\delta}$ and Abed Qaddoumi$^{\rho}$\\
  $^{\dagger}$Boston University \hspace{0.3cm} $^{\Delta}$University of Pittsburgh \hspace{0.3cm} $^{\delta}$ETH Zurich \hspace{0.3cm} $^{\rho}$Independent \\
  \texttt{najoung@bu.edu}  \hspace{0.3cm}
  \texttt{jatin.khilnani@pitt.edu}  \hspace{0.3cm} \\
  \texttt{awarstadt@ethz.ch}   \hspace{0.3cm}
  \texttt{abdulrahim.qaddoumi@gmail.com}
  }

\begin{document}
\maketitle
\begin{abstract}
We propose \textit{reconstruction probing}, a new analysis method for contextualized representations based on reconstruction probabilities in masked language models (MLMs). This method relies on comparing the reconstruction probabilities of tokens in a given sequence when conditioned on the representation of a single token that has been fully contextualized and when conditioned on only the decontextualized lexical prior of the model. This comparison can be understood as quantifying the contribution of contextualization towards reconstruction---the difference in the reconstruction probabilities can only be attributed to the representational change of the single token induced by contextualization. We apply this analysis to three MLMs and find that contextualization boosts reconstructabilty of tokens that are close to the token being reconstructed in terms of linear and syntactic distance. Furthermore, we extend our analysis to finer-grained decomposition of contextualized representations, and we find that these boosts are largely attributable to static and positional embeddings at the input layer.
\end{abstract}

{\let\thefootnote\relax\footnotetext{$^{\dagger,\Delta,\delta,\rho}$Work partially done at New York University.}}

\section{Introduction}
Model building in contemporary Natural Language Processing usually starts with a neural network pretrained on the objective of context reconstruction (``language modeling''). Contextualized representations of complex linguistic expressions from such models have been shown to encode rich lexical and structural information \citep{tenney2019you,rogers2020primer}, making these models an effective starting point for downstream applications.

Recent work on \newterm{probing} pretrained language models aims to tease apart the kind of linguistic information encoded by these models, and quantify the degree to which their representations satisfactorily align with our understanding of human language (see \citealt{belinkov2022probing} for a review). The methodology employed in such work varies widely, from supervised classifiers targeting specific linguistic properties of interest (\citealt{ettinger2016probing,giulianelli2018hood,tenney2019bert,conia2022probing}), similarity-based analyses \citep{gari2021let,lepori2020picking}, cloze-type tests \citep{goldberg2019assessing,pandit2021probing}, and causal intervention-based methods \citep{vig2020causal,elazar2021amnesic,geiger2021causal}. This methodological diversity is beneficial given the high variability of conclusions that can be drawn from a study using a single method \citep{warstadt2019investigating}---converging evidence is necessary for a more general picture.

We contribute to this line of research through developing a new analysis method that we name \newterm{reconstruction probing}, which primarily relies on token probabilities obtained from context reconstruction, applicable to models pretrained on objectives of this kind.\footnote{\url{https://github.com/najoungkim/mlm-reconstruction}} Our method is characterized by two core properties. First, it is causal: rather than asking `what features can we extract from the contextualized representations?', we ask `what effect does contextual information have on the model predictions?' through intervention at the input level. Second, our method is behavioral: it relies on the original task that the model was already trained to perform. This obviates the need to train specialized probes, which can be difficult to interpret due to the added confound of task-specific supervision.

\begin{figure*}[t]
    \centering
    \includegraphics[width=1.7\columnwidth]{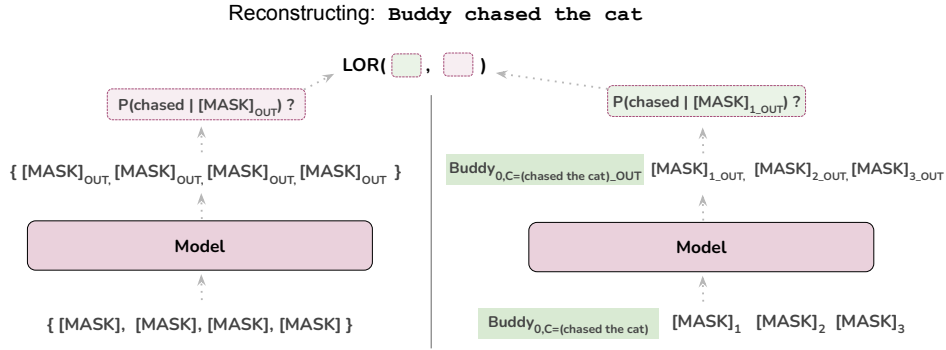}
    \caption{(Left) How the probability of \textit{chased} from only the lexical priors of the model is obtained. The input to the model is a sequence of masked tokens of the same length as the original sentence, without any positional embeddings. (Right) How the probability of \textit{chased} given a fully contextualized representation of the token \textit{Buddy} is computed (see Figure~\ref{fig:diagram} for more details). The reconstruction probabilities from (Left) and (Right) are compared using log odds ratio (LOR; Eq.~\ref{eq:lor}).}
    \label{fig:overview-diagram}
\end{figure*}

Our method aims to probe how much information the contextualized representation of a \textbf{single} token contains about the other tokens that co-occur with it in a given sequence in masked language models. Our approach is to measure the difference between the reconstruction probability of a co-occurring token in the sequence given the full contextualized representation being probed, and the reconstruction probability of the same co-occurring token only from the lexical priors of the model. This method can be generalized to compare two arbitrary representations where one representation is expected to contain strictly more features than the other (e.g., a static embedding of a token vs.~an embedding of the same token created by summing the static embedding and its positional embedding in context). Any difference between the reconstruction probabilities can be attributed to the presence/absence of those features. 

Using this method, we find that the contextualized representation of a token contains more information about tokens that are closer in terms of linear and syntactic distance, but do not necessarily encode identities of those tokens. A follow-up analysis that decomposes contextualized representations furthermore shows that the gains in reconstructability we find are largely attributable to static and positional embeddings at the input layer.

\section{Proposed Approach}

Pretrained Transformer models such as BERT \citep{devlin2019bert} learn to construct contextual representations through context reconstruction objectives like masked language modeling (MLM; e.g., predicting the token in place of \texttt{[MASK]} in `The \texttt{[MASK]} sat on the mat'). The classifier's prediction is based entirely on the \texttt{[MASK]} token, meaning these representations are optimized to contain information about other tokens of the sequence and the position of the token itself insofar as this information can help to resolve the identity of the \texttt{[MASK]} token. Our approach aims to quantify how much the contextualization of the \texttt{[MASK]} token contributes to changing the MLM predictions. 

\subsection{Metric}
We operationalize \newterm{contextual informativeness} of a token representation as its contribution to predicting other tokens in the same sequence---i.e., the contribution to the MLM probability, or \newterm{reconstruction probability}.  We quantify the contribution of a more informative token representation $j^{++}$ towards reconstructing a different token $i$, by comparing the reconstruction probability $P(i|j^{++})$ to the reconstruction probability of $i$ given a less informative token representation $j$, $P(i|j)$.   

For example, you can obtain the contextualized representation of \textit{Buddy} in the input sequence \textit{Buddy chased Cookie} by passing this through a model. If \textit{Buddy$_{contextual}$} encodes information helpful for predicting \textit{chased}, the masked language modeling probability $P(\textit{chased}|R[\texttt{[MASK]}_1, $\textit{Buddy}$_{contextual}])$ would be higher than $P(\textit{chased}|R[\texttt{[MASK]}, \emptyset])$---the lexical prior of the model for \textit{chased}.\footnote{We use R[\texttt{[MASK]}$_{pos}$, \token{source}] to refer to the representation of the \texttt{[MASK]} token at position $pos$ at the output layer of the model, which is the input to the final classifier that produces the probability distribution for masked token prediction. See Section \ref{subsec:recon-prob-calculation} for a full description of how we compute reconstruction probabilities.} The difference between these probabilities is measured in terms of the log odds ratio given the base reconstruction probability $q$ (predicting from less context) and the contextualized reconstruction probability $p$ (predicting from more context): 

\begin{figure*}[t]
    \centering
    \includegraphics[width=1.7\columnwidth]{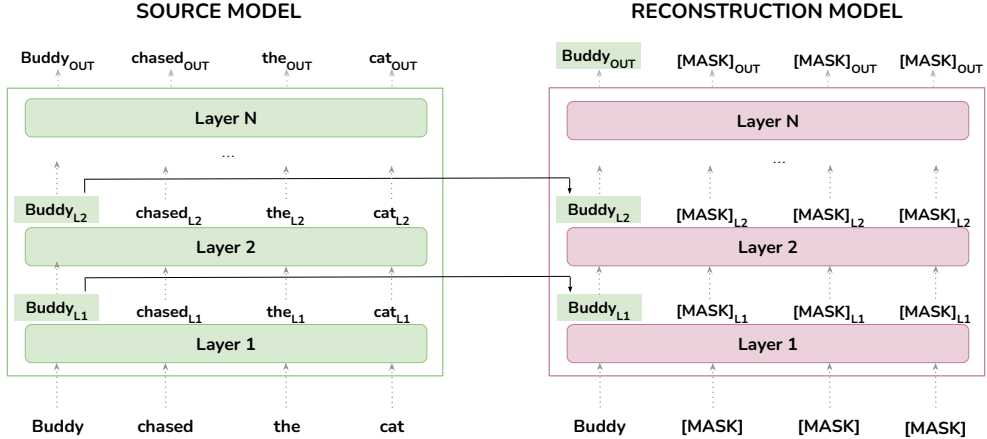}
    \caption{Diagram of the fully contextualized reconstruction setting, providing more details about how the right hand side of Figure~\ref{fig:overview-diagram} is implemented.}
    \label{fig:diagram}
\end{figure*}

\begin{equation}
    \label{eq:lor}
    \texttt{LOR}(p, q) = \ln \left (\frac{p/(1-p)}{q/(1-q)} \right)
\end{equation}

\noindent The probabilities $p$ and $q$ are defined with respect to \token{source} and \token{reconstruction} (shortened as \token{recon}) tokens. \token{Source} tokens refer to tokens that are revealed to the model at prediction time (e.g., \textit{Buddy} in the running example). \token{Recon} tokens are tokens in the original sequence the model is asked to predict (e.g., \textit{chased} in the running example). In obtaining probabilities $p$ and $q$, the \token{recon} tokens are replaced with \texttt{[MASK]} tokens, only leaving the \token{source} token revealed to the model (more detailed description is given in Section~\ref{subsec:recon-prob-calculation}). MLM probability of the token in the original sequence is computed for each \texttt{[MASK]} token in the probe input---for instance, for \textit{Buddy}$_{contextual}$ [MASK] [MASK], we compute the probability of \textit{chased} at position 1 given this sequence, and \textit{Cookie} at position 2 given this sequence. We compute Eq.~\ref{eq:lor} for every pair of tokens ($t_i, t_j$) in a given sequence, where $t_i$ is \token{source} and $t_j$ is \token{recon}. This value represents the degree of change in the probability of the reconstruction token $t_j$ induced by the contextualization of the source token $t_i$. 

\subsection{Obtaining the Reconstruction Probabilities}
\label{subsec:recon-prob-calculation}

We use the metric proposed above to gauge the contribution of a contextualized representation of a single token in reconstructing its context, over and above the lexical prior (i.e., completely context-independent) of the model as illustrated in Figure~\ref{fig:overview-diagram}. We describe below how the reconstruction probabilities from a fully contextualized representation and from the lexical prior of the model are obtained.

\paragraph{Fully Contextualized} To obtain a fully contextualized representation of a token in a particular sequence (e.g., \textit{Buddy chased Cookie}), we first pass the original, unmasked sequence of tokens through a masked language model. Here, we save each contextualized token representation at each layer of the model (e.g., \textit{Buddy}$_{L1}$, \textit{Buddy}$_{L2}$, \dots, \textit{Buddy}$_{Lm}$ where $m$ is the number of layers).  Then, we create $n$ ($n=|seq|$) versions of the input sequence where only a single token is revealed (\textit{Buddy} \texttt{[MASK]} \texttt{[MASK]}, \texttt{[MASK]} \textit{chased} \texttt{[MASK]}, \texttt{[MASK]} \texttt{[MASK]} \textit{Cookie}). We pass each sequence through the same masked language model, but at each layer, we replace the representation of the unmasked token with the stored contextualized representation of that token (see Figure~\ref{fig:diagram} for an illustration). Then, in order for the masked language modeling head to predict each \texttt{[MASK]} token in the sequence, it can only rely on the information from the representation of the single unmasked token (\token{source}), where the \token{source} token representation is contextualized with respect to the original, fully unmasked sequence. For each \texttt{[MASK]} token in the sequence, we take the probability of the token in the same position in the original sequence as the reconstruction probability. For example, $P(\textit{chased}|C[\texttt{[MASK]}_1, $\textit{Buddy}$_{contextual}])$ and $P(\textit{Cookie}|C[\texttt{[MASK]}_2, $\textit{Buddy}$_{contextual}])$ are the reconstruction probabilities of \textit{chased} and \textit{Cookie}, respectively, given the representation of fully contextualized \textit{Buddy}.  

\paragraph{Lexical Prior Only Baseline} 
We pass through a fully masked version of the input sequence as above, but do not add the positional embeddings at the input layer. The reconstruction probability that we obtain here corresponds to the probability of predicting the token in the original sequence in the absence of any lexical information \textit{and} positional information. We expect this probability to reflect a general prior of the model over the vocabulary, for instance based on frequency in the training corpus.

\section{Experiment Setup}

\begin{figure*}[h]
    \centering
    \begin{subfigure}[b]{0.33\textwidth}
         \centering
         \includegraphics[width=\textwidth]{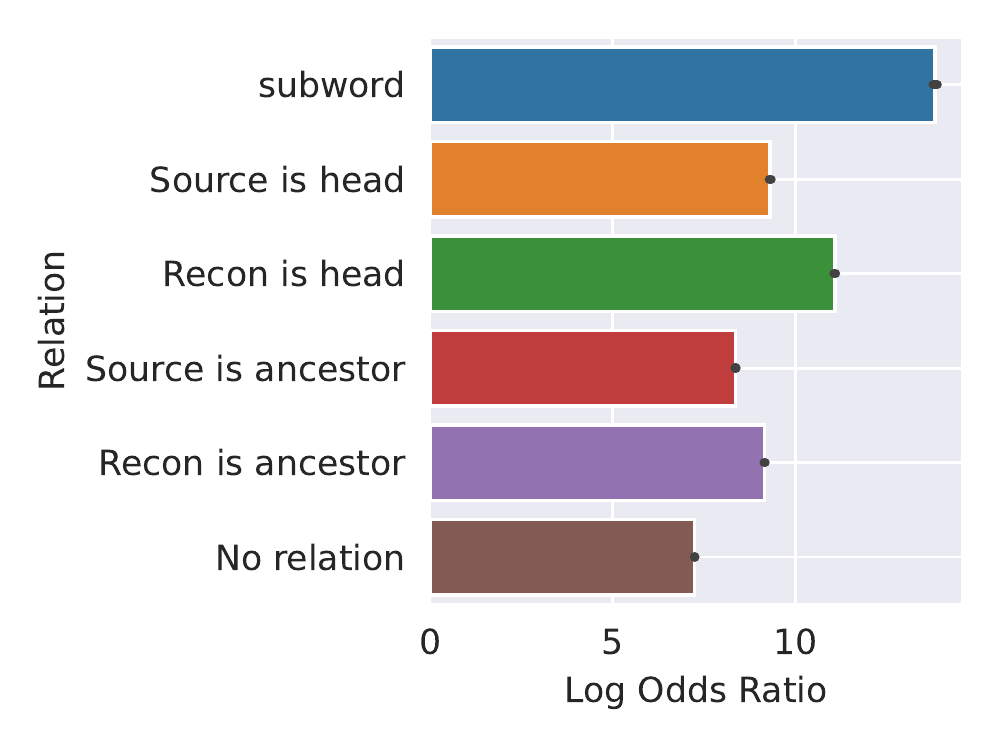}         \caption{BERT}
    \end{subfigure}%
    \begin{subfigure}[b]{0.33\textwidth}
         \centering
         \includegraphics[width=\textwidth]{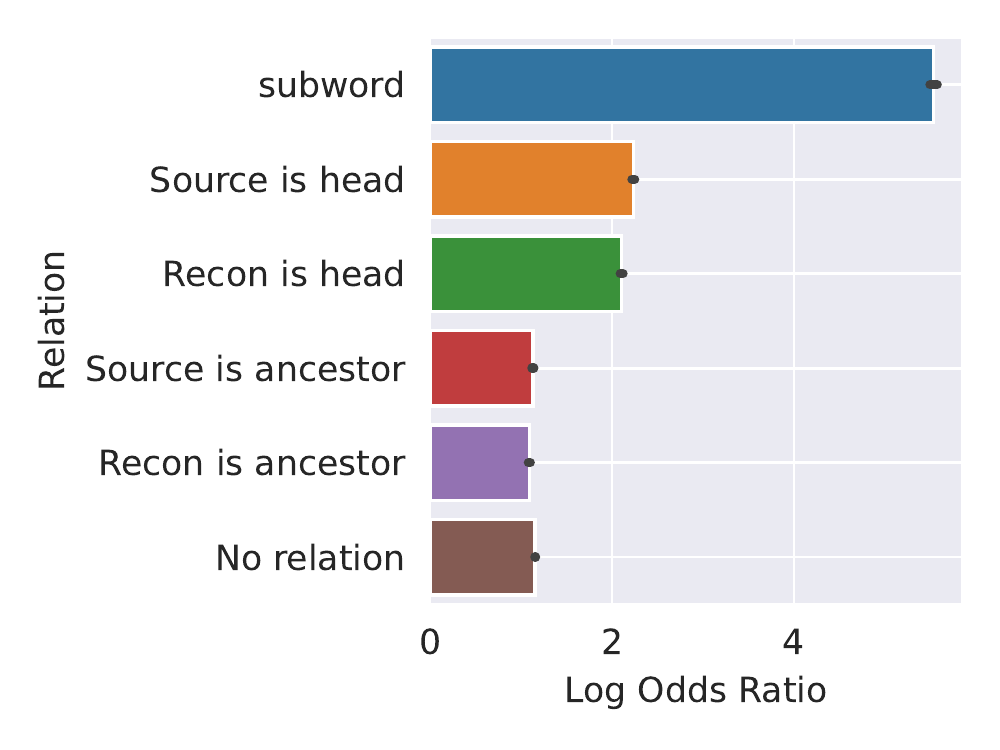}         \caption{RoBERTa}
    \end{subfigure}%
    \begin{subfigure}[b]{0.33\textwidth}
         \centering
         \includegraphics[width=\textwidth]{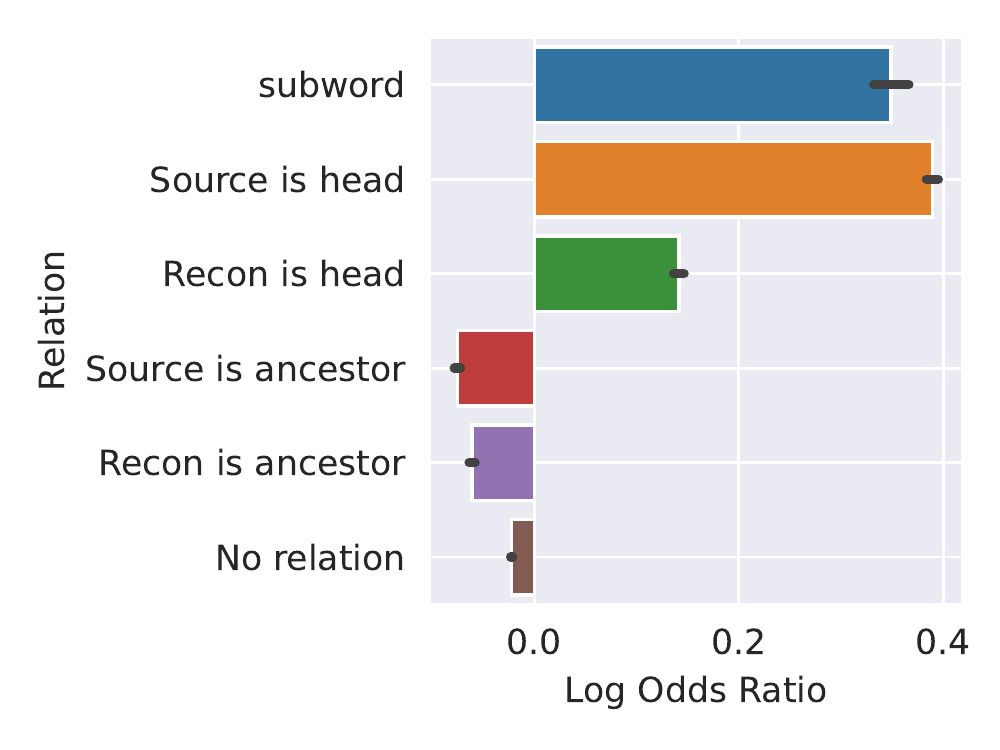}         \caption{DistilBERT}
    \end{subfigure}
    
    \caption{Reconstructibility boost by syntactic relation measured by log odds ratio.}
    \label{fig:relation}
\end{figure*}

\subsection{Models}
We analyzed three Transformer-based masked language models widely used for obtaining contextualized representations: BERT \citep{devlin2019bert}, RoBERTa \citep{liu2019roberta} and DistilBERT \citep{sanh2019distilbert}. All three models were pretrained using the masked language modeling objective (BERT and DistilBERT, also on Next Sentence Prediction), and they all share a basic architecture. RoBERTa has been shown to outperform BERT on many downstream NLP tasks as well as showing stronger performance on intrinstic linguistic evaluation (cite). DistilBERT is a more compact version of BERT obtained throug knowledge distillation, in which a student model is trained to mimic the behavior of a larger teacher model. DistilBERT has been claimed to retain much of the downstream task performance of BERT despite being substantially smaller \citep{sanh2019distilbert},
and has been shown to be highly similar to BERT in terms of constituency trees that can be reconstructed from linear probes \citep{arps2022probing}.

\subsection{Data}
We used sentences from the Multi-Genre Natural Language Inference (MNLI; \citealt{williams2018broad}) dataset for this analysis. We selected MNLI because it contains sentences of varying lengths from a range of domains, and is not a part of the pretraining data of the models we are probing. We then sampled 10K premise sentences from the non-spoken genres of the dataset (i.e., excluding \textsc{Telephone} and \textsc{Face-to-Face}). We excluded spoken data as it is less typical of the data domain the models were trained on, and we excluded hypothesis sentences because they were generated by crowdworkers given the naturally-occurring premises.

\subsection{Procedure}
For each of the 10K sentences, we created two different sets of probe inputs as illustrated in Figure~\ref{fig:overview-diagram}. We passed the probe inputs to the models to obtain the two different reconstruction probabilities (from lexical prior only vs. from a fully contextualized source token) of each of the tokens in the input, as described in Section~\ref{subsec:recon-prob-calculation}. Finally, we computed the log odds ratios between the two reconstruction probabilities using Eq.~\ref{eq:lor} to quantify the contribution of contextualization for all possible (\token{source}, \token{recon}) token pairs in the original sentence.

\begin{figure*}[h]
    \centering
    \begin{subfigure}[b]{0.27\textwidth}
         \centering
         \includegraphics[width=\textwidth]{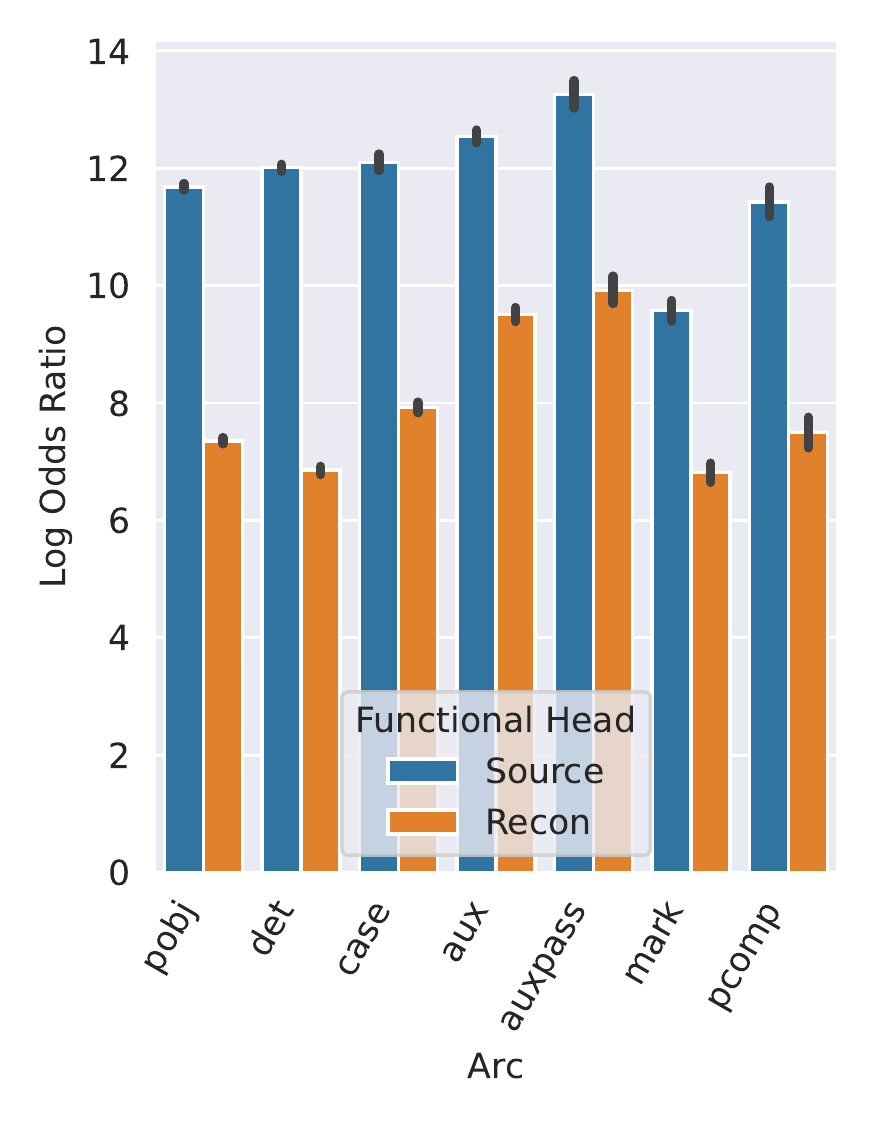}
        \caption{BERT}
    \end{subfigure}%
    \begin{subfigure}[b]{0.27\textwidth}
         \centering
         \includegraphics[width=\textwidth]{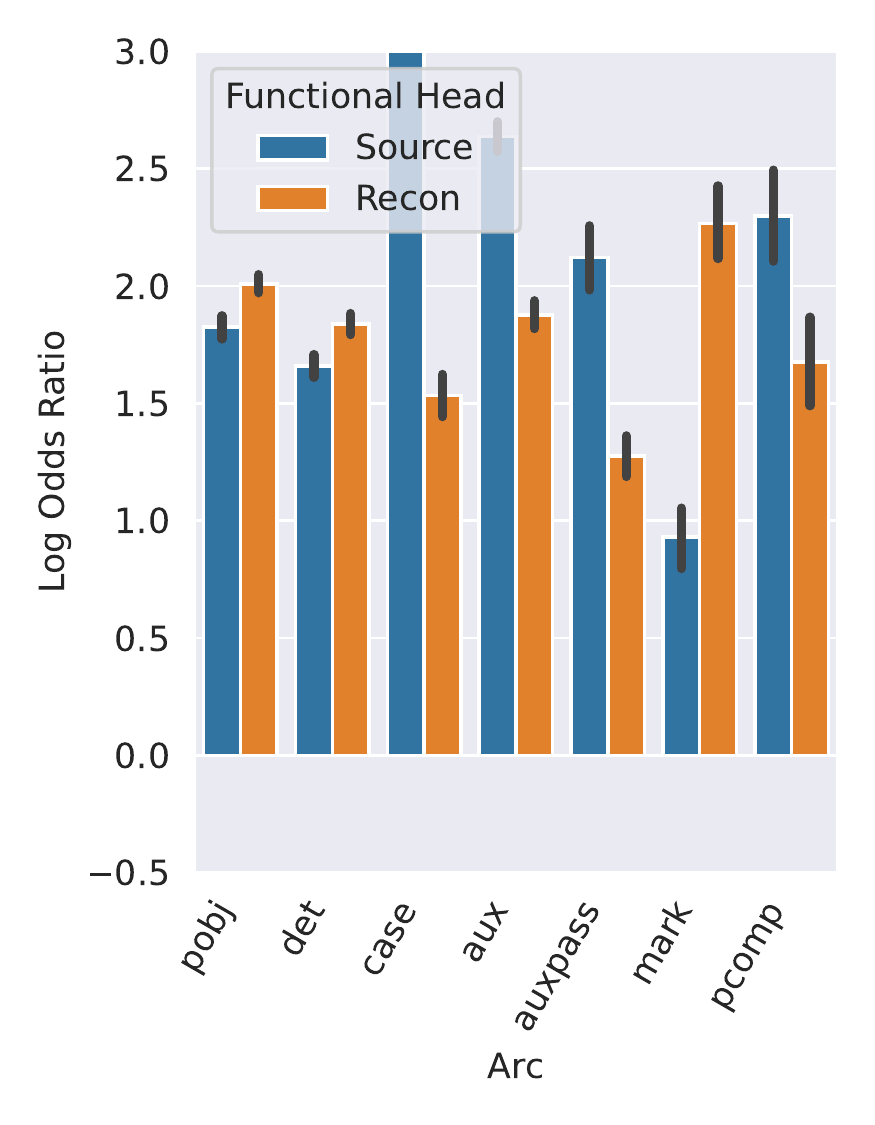}
         \caption{RoBERTa}
    \end{subfigure}%
    \begin{subfigure}[b]{0.27\textwidth}
         \centering
         \includegraphics[width=\textwidth]{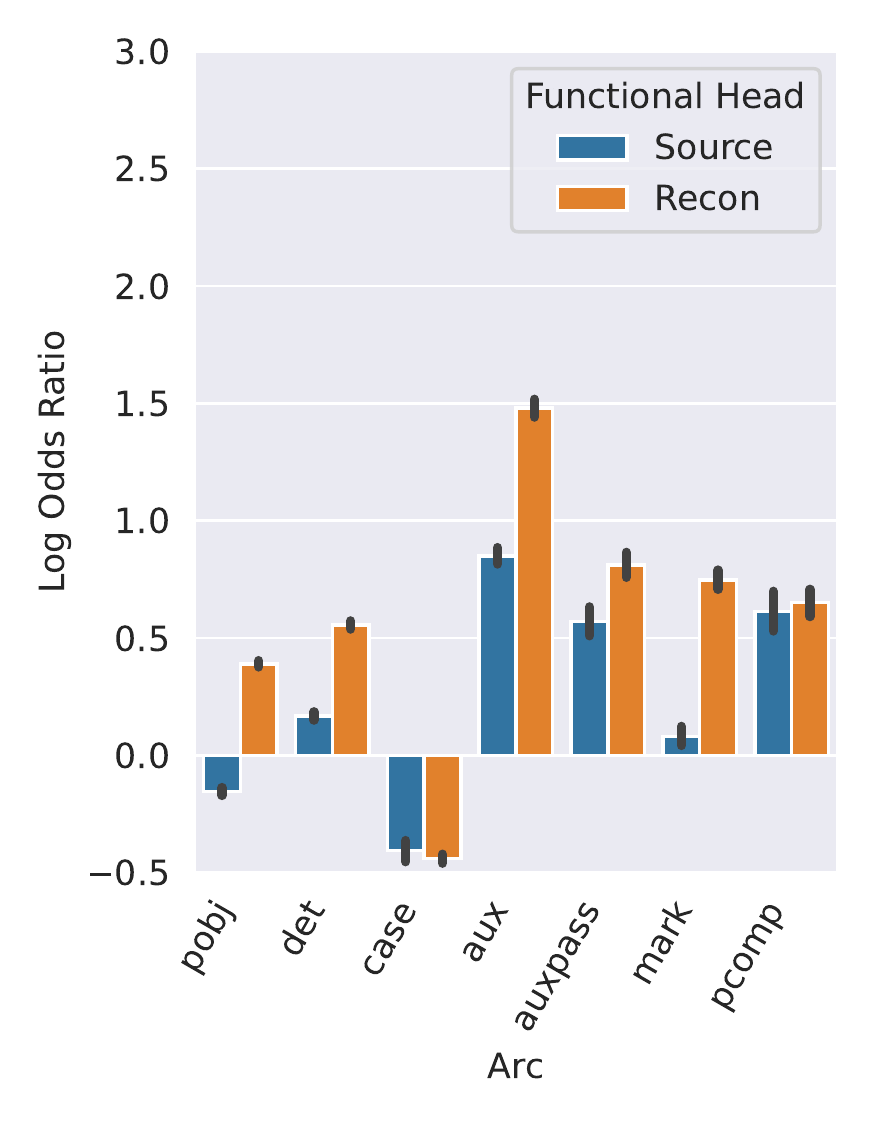}
          \caption{DistilBERT}
    \end{subfigure}

    \caption{Reconstructibility boost (log-odds ratio with vs. without source) broken down by the functional relation between a functional head and a content-word dependent.}
    \label{fig:functional_head}
\end{figure*}

\section{Analyses}

\subsection{Is Token Identity Exactly Recoverable from Contextualized Representations?}
The \token{recon} token is among the top 10 MLM predictions of the model only a small percent of the time (BERT: 22.1\%. RoBERTa: 7.9\%, DistilBERT: 8.2\%), even though the \token{source} token provided to the model has been contextualized with all co-occurring tokens revealed. This observation suggests that the information encoded in the contextualized representations are a degree more abstract than directly encoding the identities of co-occurring tokens in the same sequence. This is in line with \citeauthor{klafka2020spying}'s (\citeyear{klafka2020spying}) finding that the features of co-occurring tokens rather than their identities are often more recoverable from the contextual representations.

\subsection{Is Reconstructability Greater when \token{Source/Recon} Tokens are in a Syntactic Relation?}
\label{subsec:aggregate-relation-analysis}

\begin{figure}[h]
    \centering
    \includegraphics[width=1\columnwidth]{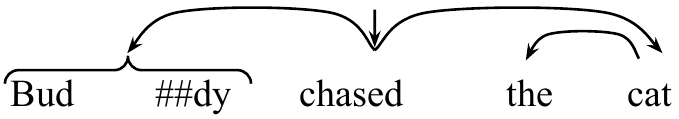}
    \caption{The dependency parse of the sentence \textit{Buddy chased the cat}.}
    \label{fig:dependency-arc}
\end{figure}

We hypothesize that the contextual information in an embedding should disproportionally reflect the syntactic neighbors of the word. To test this hypothesis, we partition reconstructability scores based on the syntactic relation between the \token{source} and \token{recon} tokens as follows:
\begin{enumerate*}[label=(\arabic*)]
    \item \textbf{\token{Source/recon} is head}: Cases where there is a single dependency arc between two tokens, the closest dependency relation possible with the exception of subword tokens. Reconstructing \textit{cat} from \textit{chased} in Figure~\ref{fig:dependency-arc} would be a case of \token{source} is head, and \textit{chased} from \textit{cat} would be of \token{recon} is head.
    \item \textbf{\token{Source/recon} is ancestor}: Cases where there is more than one dependency arc connecting the two tokens. Reconstructing \textit{the} from \textit{chased}  would be a case of \token{source} is ancestor, and \textit{chased} from \textit{the} would be of \token{recon} is ancestor.
    \item \textbf{subword}: \token{source/recon} tokens are subwords of the same lexical item. \textit{Bud} and \textit{\#\#dy} is an example.
    \item \textbf{No relation}: None of the above relations holds. For example, tokens \textit{Bud} and \textit{the} are not in a dependency relation.
\end{enumerate*}

\noindent Our results in Figure \ref{fig:relation} confirm our hypothesis. In general, we find that the degree to which contextual information improves reconstruction depends on the existence of a syntactic relation between the \token{source} and \token{recon} as expected. In all models, tokens in a subword or head-dependent relation are more reconstructable from each other compared to tokens with no relation. Furthermore, among tokens that are in a dependency relation, the closer the relation, the higher the reconstruction boost: reconstruction boost is the greatest for tokens in a subword relation, then for tokens in a head-dependent relation, and then for tokens in ancestor-descendant relation. These trends were consistent across all models we evaluate, with the exception of DistilBERT where reconstruction boost when \token{source} is head was greater than tokens in a subword relation. The models showed more variation in whether ancestor relations boosted reconstructability significantly. While tokens in an ancestor-descendant relation (excluding direct dependents) were more reconstructable than tokens not in a dependency relation in BERT, this was not the case for RoBERTa and DistilBERT. We also did not find a large or consistent effect of whether the \token{source} token or the \token{recon} token is the ancestor (including direct head-dependent relations). Thus we cannot conclude that ancestors tend to contain more information about descendants than vice-versa.

\subsection{Finer-Grained Syntactic Properties}
\label{subsec:finegrained-relation-analysis}

In the next set of analyses, we study how fine-grained syntactic properties of the words affect reconstructability, focusing on cases where there is a syntactic relation between \token{source} and \token{recon}. 

\paragraph{Dependency Relations} One natural way to break down the results is by the label of the dependency relation that holds between \token{source} and \token{recon} when such a relation exists. However, we did not find an overarching trends; results were generally idiosyncratic to models although boost for token pairs in \textsc{root} and \textsc{prt} (particle) relations was high across all models. See Appendix~\ref{app:dep-arc} for full results.

\paragraph{Functional Relations} 

Next, we zoom in on relations between functional heads and their content-word dependents (Figure \ref{fig:functional_head}). Table \ref{tab:functional} lists all the dependency arcs we use to identify functional heads.\footnote{While function words are typically considered heads of content words in linguistic theory, the opposite is often true in Universal Dependencies (UD ver.~2: \url{https://universaldependencies.org/en/index.html}). 
} 
First, we find that reconstructability is generally high for these pairs. Second, auxiliary-verb relations are associated with particularly high reconstructability for all models. One possible explanation for this finding is the fact that there is always morphological agreement between auxiliaries and verbs, unlike most other functional relations. Third, among functional relations, reconstructability is always lowest for complementizer-verb relations (labeled \emph{mark}). We speculate that the complementizer might encode contextual information about the entire complement clause, which often includes many more content words than just the head verb.

\begin{table}[h]
    \centering
    \resizebox{\columnwidth}{!}{%
    \begin{tabular}{lll}
    \toprule
    \bf Relation & \bf FW is ... & \bf Example\\\midrule
    aux & Dependent & \emph{The dog \ul{is} \ul{sleeping}.}\\
    auxpass & Dependent &  \emph{The dog \ul{was} \ul{taken} out.}\\
    case & Dependent & \emph{The \ul{dog} \ul{'s} bone is gone.}\\
    det & Dependent &  \emph{\ul{The} \ul{dog} barked.}\\
    mark & Dependent & \emph{I think \ul{that} the dog \ul{ate}.}\\
    pcomp & Head & \emph{I dream \ul{about} dogs \ul{playing}.} \\
    pobj & Head & \emph{I played \ul{with} the \ul{dog}.}\\\bottomrule
    \end{tabular}
    }
    \caption{Dependency arcs denoting functional relations. \textbf{FW is ...} indicates whether the function word is considered the head or the dependent in UD.}
    \label{tab:functional}
\end{table}

We hypothesized that functional heads should encode more information about their dependents in context than vice-versa. Our rationale was that function words carry less information than content words, but their contextual representations are equal in size, leaving more space for information about the rest of the sentence.
Our findings are ambiguous with respect to this hypothesis. Results from BERT support the hypothesis for all relations. On the other hand, no consistent asymmetry was observed for RoBERTa, and for DistilBERT, the observed pattern mostly contradicts our hypothesis. The large difference between BERT and DistilBERT results goes against prior results that suggest that the syntactic trees recoverable from these two models are highly similar \citep{arps2022probing}.

\subsection{Linear and Structural Distance}
We also hypothesized that the distance between two tokens (both in linear and structural terms) would affect reconstruction. Linear distance is the difference between the linear indices of \token{source} and \token{recon}: if they are the $i^{th}$ and $j^{th}$ tokens respectively, their linear distance is $|i-j|$. Structural distance is the number of arcs in the directed path between \token{source} and \token{recon} tokens (if there is a path). For example, in Figure~\ref{fig:dependency-arc} the structural distance between \emph{the} and \emph{chased} is 2.

\begin{figure}[htbp]
    \centering
    \begin{subfigure}[b]{0.15\textwidth}
         \centering
         \includegraphics[width=\textwidth]{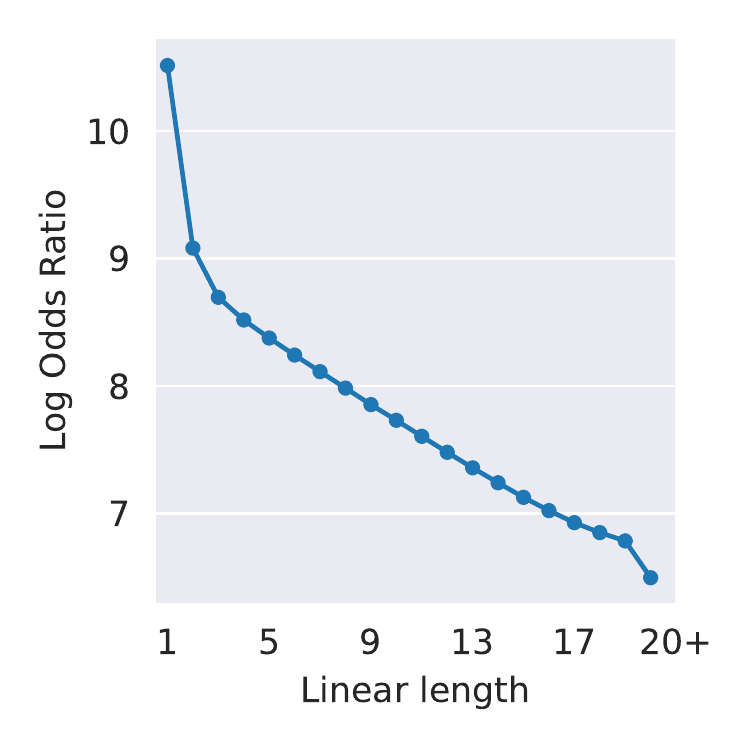}
         \includegraphics[width=\textwidth]{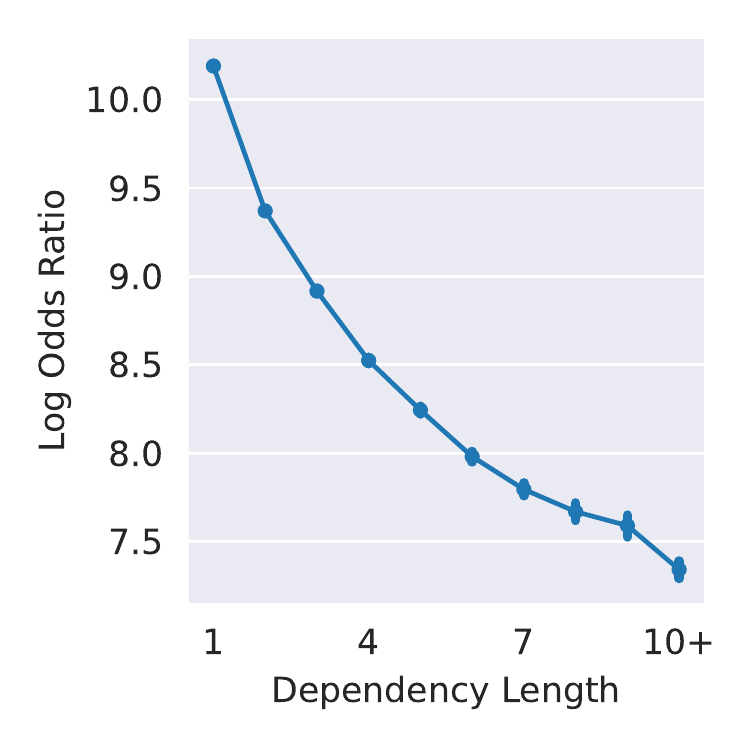}         
         \caption{BERT}
    \end{subfigure}%
    \begin{subfigure}[b]{0.15\textwidth}
         \centering
         \includegraphics[width=\textwidth]{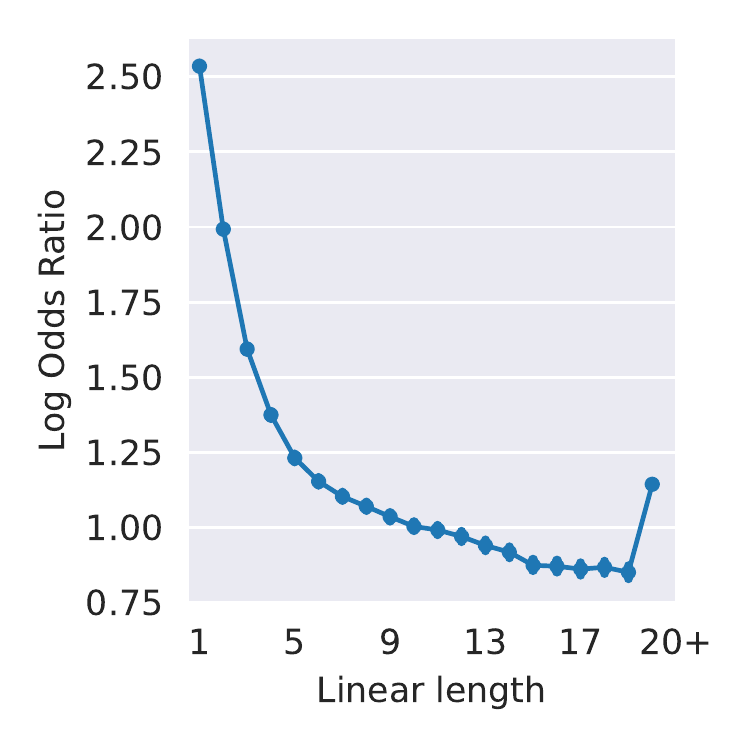}         \includegraphics[width=\textwidth]{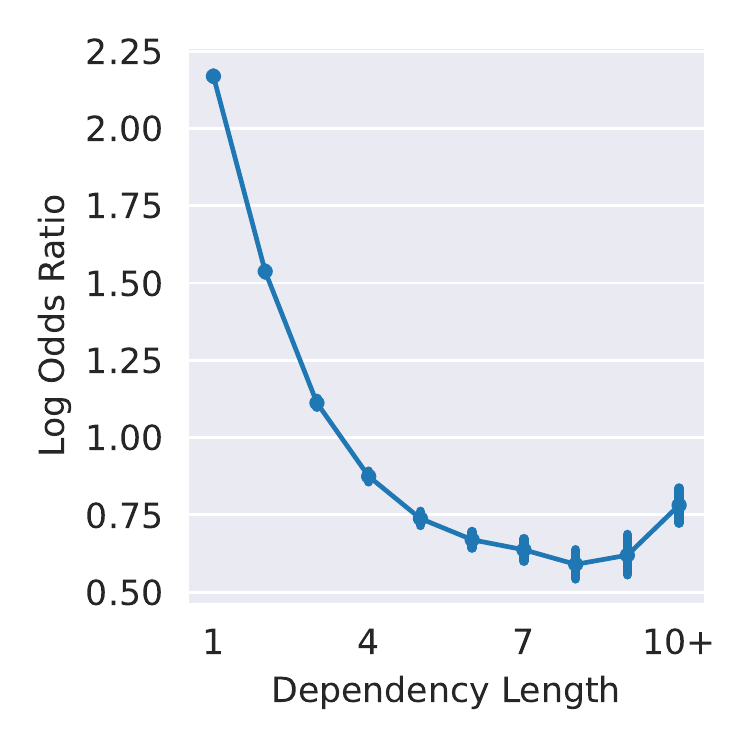}
         \caption{RoBERTa}
    \end{subfigure}%
    \begin{subfigure}[b]{0.15\textwidth}
         \centering
         \includegraphics[width=\textwidth]{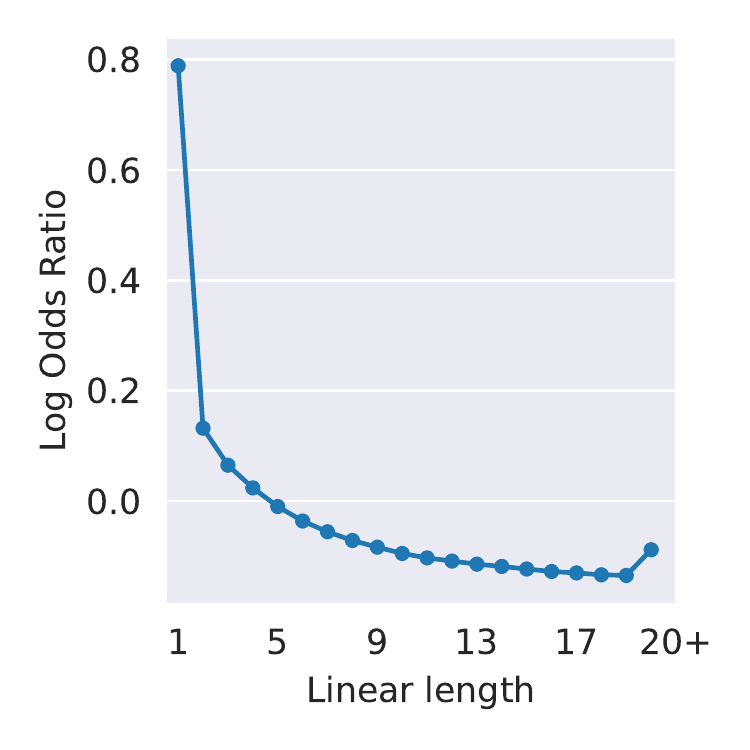}         \includegraphics[width=\textwidth]{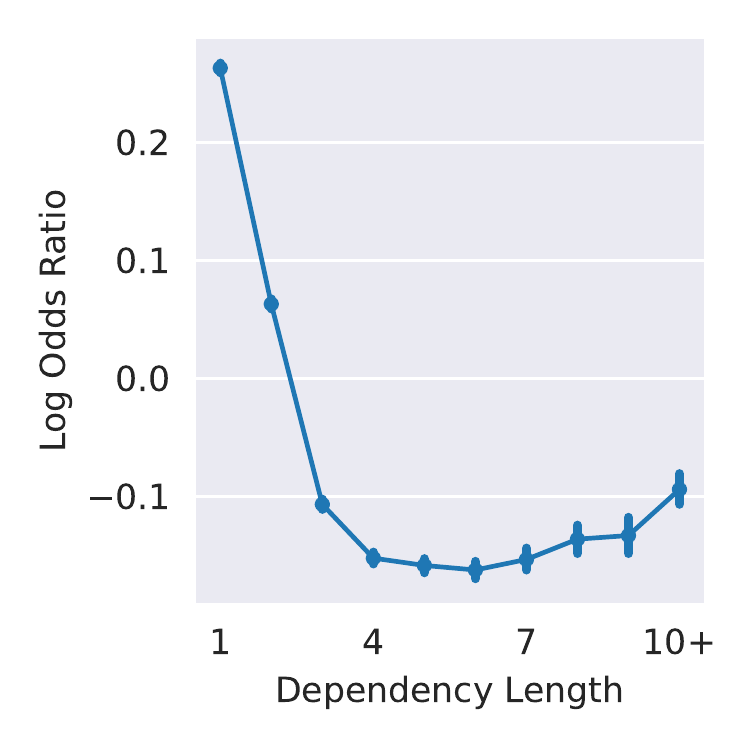}
         \caption{DistilBERT}
    \end{subfigure}
   
    \caption{Reconstructibility boost (log odds ratio) broken down by linear distance (top) and structural distance (bottom) between \token{source} and \token{recon}.}
    \label{fig:linear-distance}
\end{figure}

\begin{table*}[]
    \centering
    \resizebox{2\columnwidth}{!}{%
    \begin{tabular}{lll}
    \toprule
    \bf  & \bf Ablated sequence & \bf Probe input (\token{source} == `Buddy') \\\midrule
    Sequence to reconstruct & Buddy chased Cookie & -\\\midrule
    Fully contextualized & Buddy$_{contextual}$ chased$_{contextual}$ Cookie$_{contextual}$ & Buddy$_{contextual}$ [MASK] [MASK]  \\
    Static embedding (+position) & Buddy$_{static}$ chased$_{static}$ Cookie$_{static}$ & Buddy$_{static}$ [MASK] [MASK]\\
    Static embedding (-position) & \{Buddy$_{static}$, chased$_{static}$, Cookie$_{static}$\} & \{Buddy$_{static}$, [MASK], [MASK]\} \\
    All mask (+position) & [MASK] [MASK] [MASK] & [MASK] [MASK] [MASK] \\
    All mask (-position) (Lexical prior only) & \{[MASK], [MASK], [MASK]\} & \{[MASK], [MASK], [MASK]\} \\
    \bottomrule
    \end{tabular}
    }
    \caption{Ablated sequence and an example of an input passed through the model to obtain the output representations when \token{source} is `Buddy'. \{\} denotes an unordered set (i.e., no positional information).}
    \label{tab:ablation}
\end{table*}

\paragraph{Linear Distance} Predictably, we find that information encoded in contextualized representations is biased towards nearby tokens in linear space (Figure~\ref{fig:linear-distance}, row 1). In other words, we find that reconstructability generally decreases with increase in linear distance. For all models, the sharpest decrease is observed between 1- and 2-token distances. Beyond this, reconstructability decreases approximately linearly in BERT, and more gradually in RoBERTa and DistilBERT.

\paragraph{Structural Distance} The second row of Figure~\ref{fig:linear-distance} shows the decline in reconstructability as the number of intervening nodes in the dependency path between the tokens increases when comparing reconstruction. This trend is strictly monotonic in BERT, but there is an small increase starting from dependency depth 7 in RoBERTa and DistilBERT. Due to the high variance in the deeper depth cases, it is unclear whether this is a genuine effect of contextualization.

\section{Decomposing Contextualization}
While we examined the effect of contextualization compared to the lexical prior only baseline, our method allows for a finer-grained decomposition of the components of contextualization. In pretrained Transformer models the input representation of a token is a function of the static embedding and a (context-specific) positional embedding. Using our method, we can study the individual influence of the static embedding, positional embedding, and remaining sequence-specific contextualization (i.e., everything that happens beyond the input layer, \textit{full contextualization} henceforth). 

We create various ablated versions of a fully contextualized sequence, as shown in the \textbf{Ablated sequence} column of Table~\ref{tab:ablation}. The reconstruction probabilities from these ablated sequences allow us to probe the contribution of the various components of contextualized language models. \textbf{Fully contextualized} and \textbf{All mask (-position)} in Table~\ref{tab:ablation} correspond to the reconstruction probabilities described and compared in Section~\ref{subsec:recon-prob-calculation}, and the rest are intermediate ablations.

\begin{figure*}[h]
    \centering
    \begin{subfigure}[b]{0.27\textwidth}
         \centering
         \includegraphics[width=\textwidth]{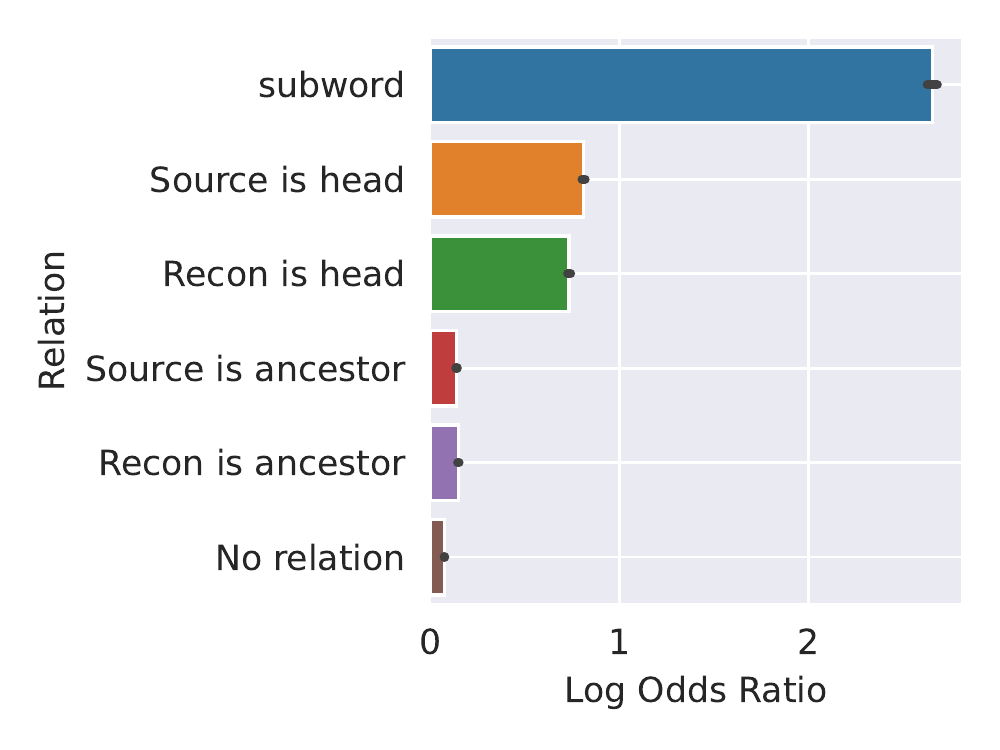}
         \includegraphics[width=\textwidth]{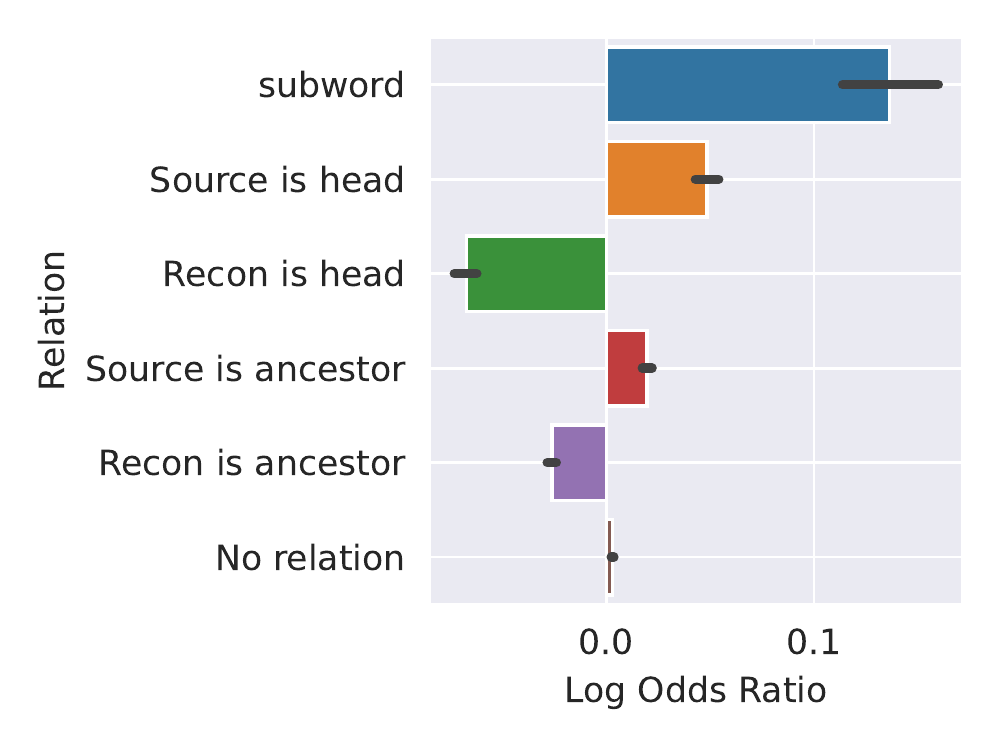}
         \caption{BERT}
    \end{subfigure}%
    \begin{subfigure}[b]{0.27\textwidth}
         \centering
         \includegraphics[width=\textwidth]{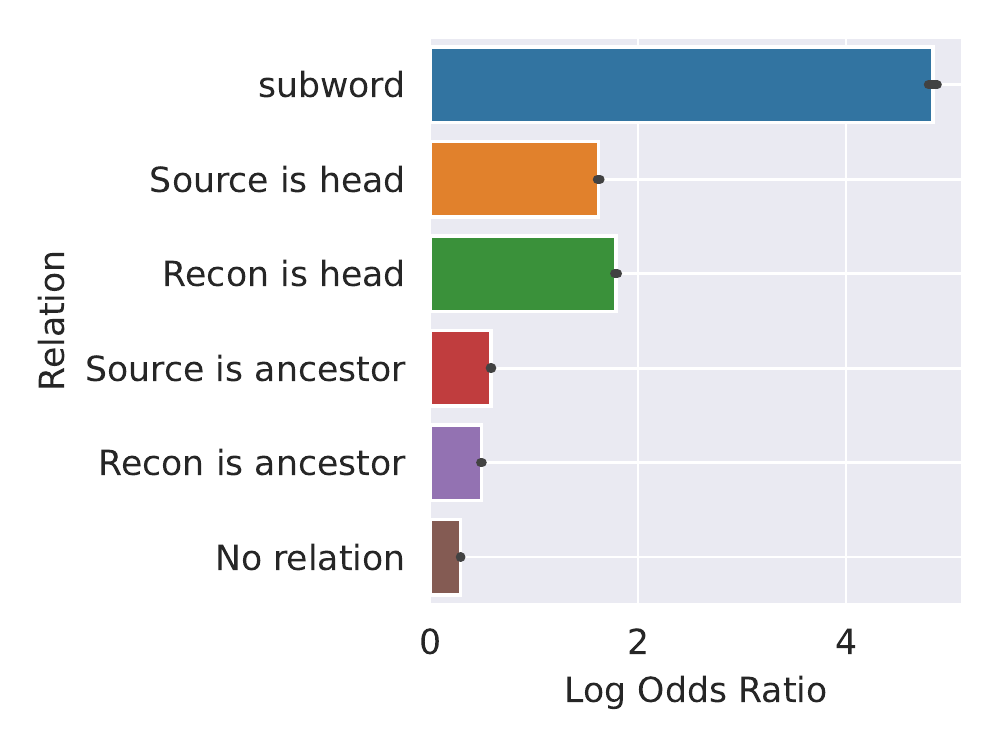}
         \includegraphics[width=\textwidth]{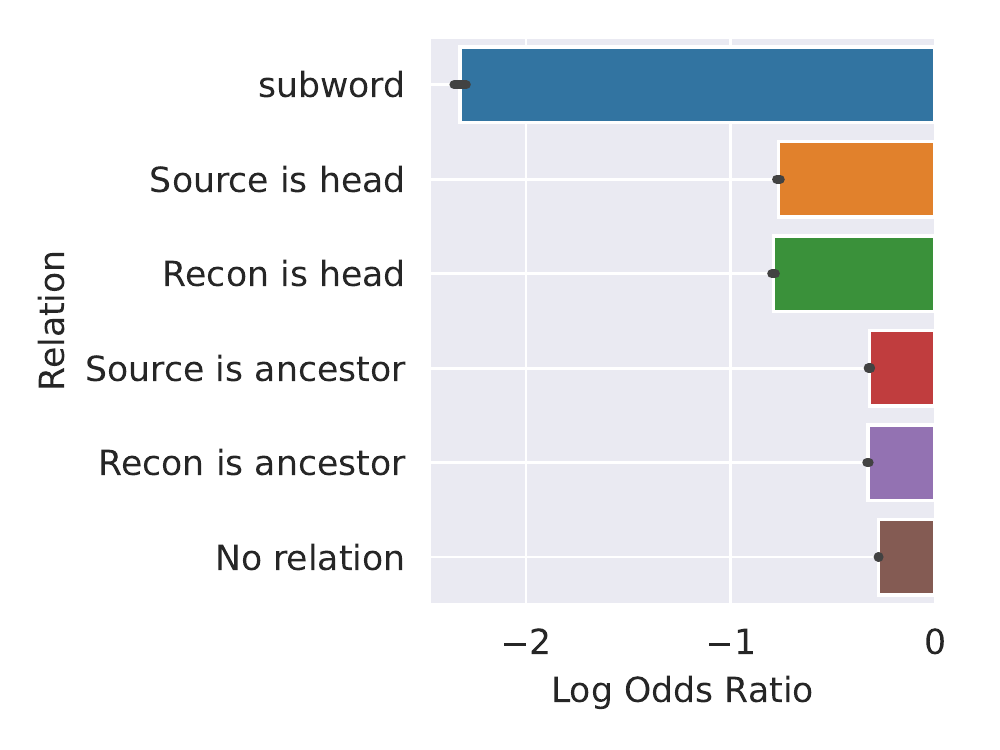}
         \caption{RoBERTa}
    \end{subfigure}%
    \begin{subfigure}[b]{0.27\textwidth}
         \centering
         \includegraphics[width=\textwidth]{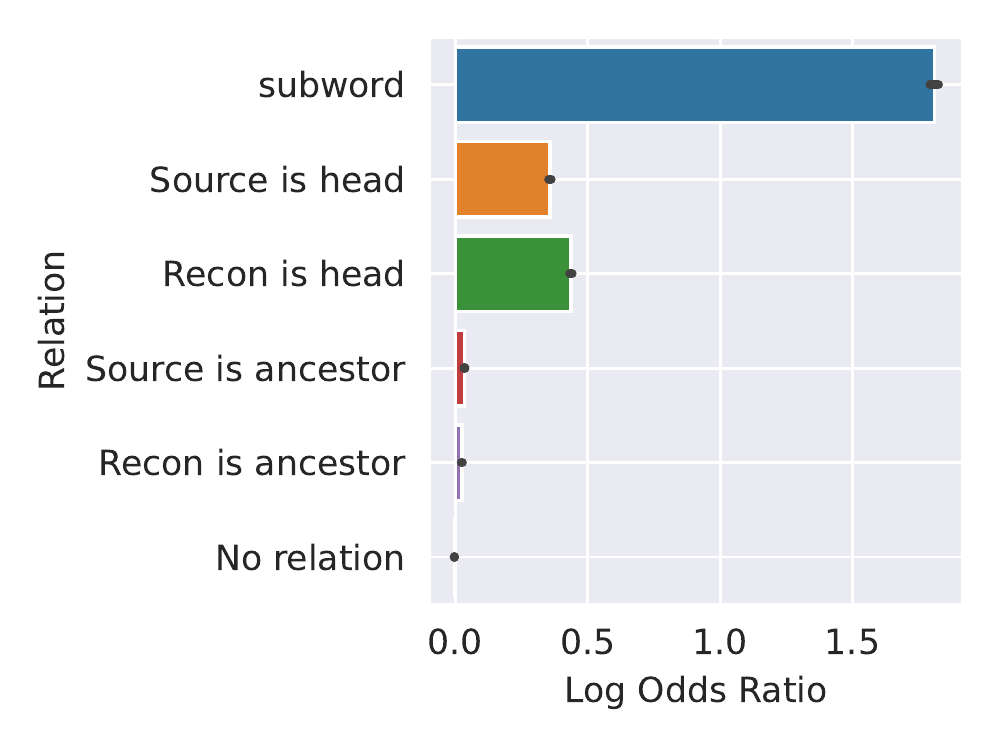}
         \includegraphics[width=\textwidth]{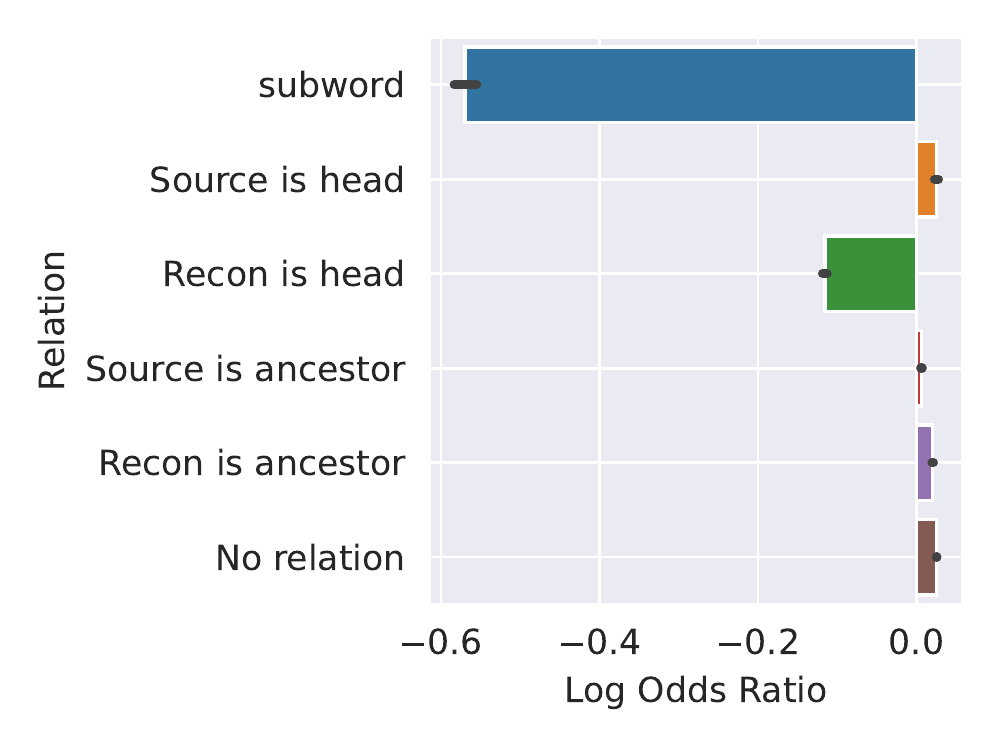}
         \caption{DistilBERT}
    \end{subfigure}
    \caption{Relative reconstructibility (log odds ratio) for static and all masked \token{source} embeddings (top) and for and fully contextualized and static \token{source} embeddings (bottom).}
    \label{fig:relation_contextual_vs_static}
\end{figure*}

\subsection{Results}
Surprisingly, we find that there is often no clear benefit to reconstruction of providing the model with the contextualized embeddings at each layer, over just providing the input embedding (static + positional embeddings) of the source token (Figure~\ref{fig:relation_contextual_vs_static}, bottom). While BERT does gain reconstructability from full contextualization for subwords and when \token{source} is an ancestor, contextualization is generally harmful or at least not helpful to reconstruction for RoBERTa and DistilBERT. This suggests that the positive reconstruction boost observed in Figure~\ref{fig:relation} may be driven largely by static embeddings, as can be seen from the gains in reconstructability in models provided with the static embeddings of the \token{source} tokens compared to models provided with only \texttt{[MASK]} tokens (Static embedding (+position) vs. All mask (+position) from Table ~\ref{tab:ablation}; Figure~\ref{fig:relation_contextual_vs_static}, top). We provide full comparisons between ablations and their interpretation in Appendix~\ref{app:decomp-analysis}.

\paragraph{When is full contextualization helpful/harmful?} To better understand the effect of full contextualization, we manually examined token pairs where the reconstruction probabilities differ the most given the static and contextual \token{source} tokens. In BERT and DistilBERT, the majority (52\% and 80\%, respectively) of the 100 most helpful scenarios of full contextualization involved reconstruction of an apostrophe in a contraction from single-character or bi-character tokens (e.g., \textit{m}, \textit{t}, \textit{re}). As the source token is highly ambiguous on its own, contextualization seems to provide additional information that these (bi)character tokens are a part of a contraction (e.g., \textit{I'm}, \textit{wasn't}, \textit{we're}. In RoBERTa, we found no interpretable pattern. 

When contextualization negatively affected reconstruction probability, it was often in cases where the \token{source} and \token{recon} formed a frequent bigram (e.g., (\textit{prix}, \textit{grand}), (\textit{according}, \textit{to}), (\textit{\#\#ritan}, \textit{pu}), (\textit{United}, \textit{States})). Since the \token{recon} token is largely predictable from static information about the \token{source} alone, contextualization seems to only dilute the signal. 

Although we find that reconstruction is often better given static embeddings than contextual ones, we reject the interpretation that contextualization is harmful to the models. This is too unlikely given overwhelming prior evidence \citep{tenney2019you} that contextualized embeddings are a big improvement over static counterparts. Rather, it is more likely that this result is an idiosyncrasy of doing context reconstruction given all masks or of the procedure for transferring the contextualized source token, both of which fall outside the setting in which these models were trained.

\section{Related Work}

Our research question is the most similar to \citet{klafka2020spying}: how much information about other tokens in the context is contained in the contextualized representation(s) of a token? While \citet{klafka2020spying} use supervised probing classifiers to predict properties of the other words in the sentence, our approach measures changes in reconstruction probability given more/less informative token representations. Our findings about better reconstructability between tokens in a syntactic dependency relation echo prior work that show sensitivity of MLMs to part-of-speech and other syntactic relations \citep{tenney2019you,goldberg2019assessing,htut2019attention,kim2021testing}. A novel finding is that some of the syntactic dependency between tokens can be traced back to \textit{static} information in the input layer, complementing dynamic layerwise analysis in works such as \citet{tenney2019bert} and \citet{jawahar2019bert}. This result aligns with \citet{futrell2019syntactic}'s observation that syntactic dependency is reflected in the corpus distribution as encoded in static embeddings.

While existing work does provide analysis of static embeddings from contextualized models \citep{bommasani2020interpreting,chronis2020bishop,sajjad2021effect}, most of this literature concerns the \textit{distillation} of static embeddings rather than isolating the contribution of static embeddings in contextualized prediction as in our work.

The current method shares similar goals with intervention-based methods such as \citet{geiger2021causal} and \citet{wu2020perturbed}, although our method examines effects of the intervention on the masked language modeling probability itself rather than separate downstream tasks. \citet{karidi2021putting} employs the most similar methodology to ours, in that they use predictions from the masked language modeling objective directly for probing. However, their primary target of analysis is the role of contextualization in word sense disambiguation.

\section{Conclusion}

We have proposed \textit{reconstruction probing}, a novel analysis method that compares reconstruction probabilities of tokens in the original sequence given different amounts of contextual information. Overall, reconstruction probing yields many intuitive results. We find that the information encoded in these representations tend to be a degree more abstract than token identities of the neighboring tokens---often, the exact identities of co-occurring tokens are not recoverable from the contexutalized representations. Instead, reconstructability is correlated with the closeness of the syntactic relation, the linear distance, and the type of syntactic relation between the \token{source} and \token{recon} tokens. These findings add converging evidence to conclusions from previous probing studies about the implicit syntactic information of contextual embeddings (\citealt{tenney2019you}). Furthermore, our method is generalizable to comparing reconstruction probabilities from any pair of representations that differ in the degree of informativeness. We extended our analysis to finer-grained decomposition of the components that constitute contextualized representations using this method, finding that most of the reconstruction gains we saw were attributable to information contained in static and positional embeddings at the input layer. This calls for deeper investigations into the role of token representations at the input layer, complementing a large body of existing work on layer-wise analysis of contextualized language models.

% Entries for the entire Anthology, followed by custom entries
\bibliography{anthology,custom}
\bibliographystyle{acl_natbib}

\appendix

\section{Dependency Relations}
\label{app:dep-arc}

\begin{figure*}[h]
    \centering
    \begin{subfigure}[b]{0.3\textwidth}
         \centering
         \includegraphics[width=\textwidth]{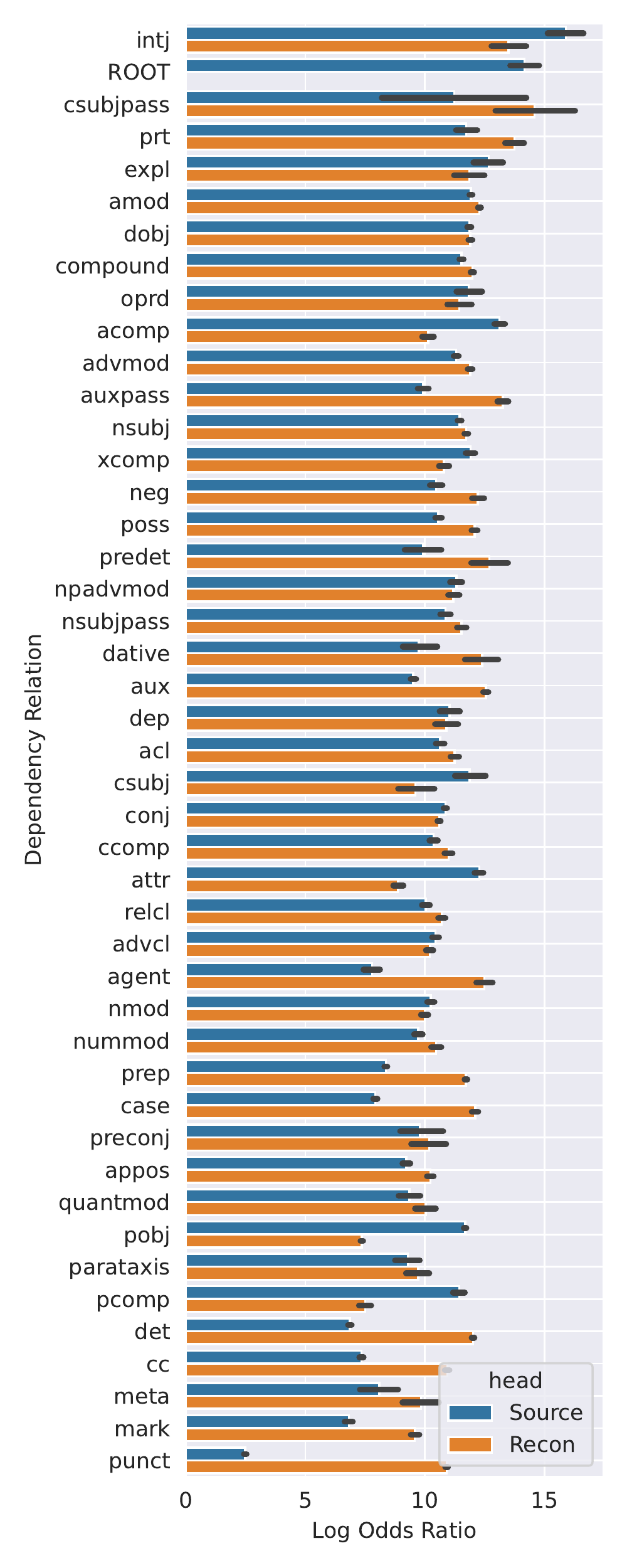}
        \caption{BERT}
    \end{subfigure}%
    \begin{subfigure}[b]{0.3\textwidth}
         \centering
         \includegraphics[width=\textwidth]{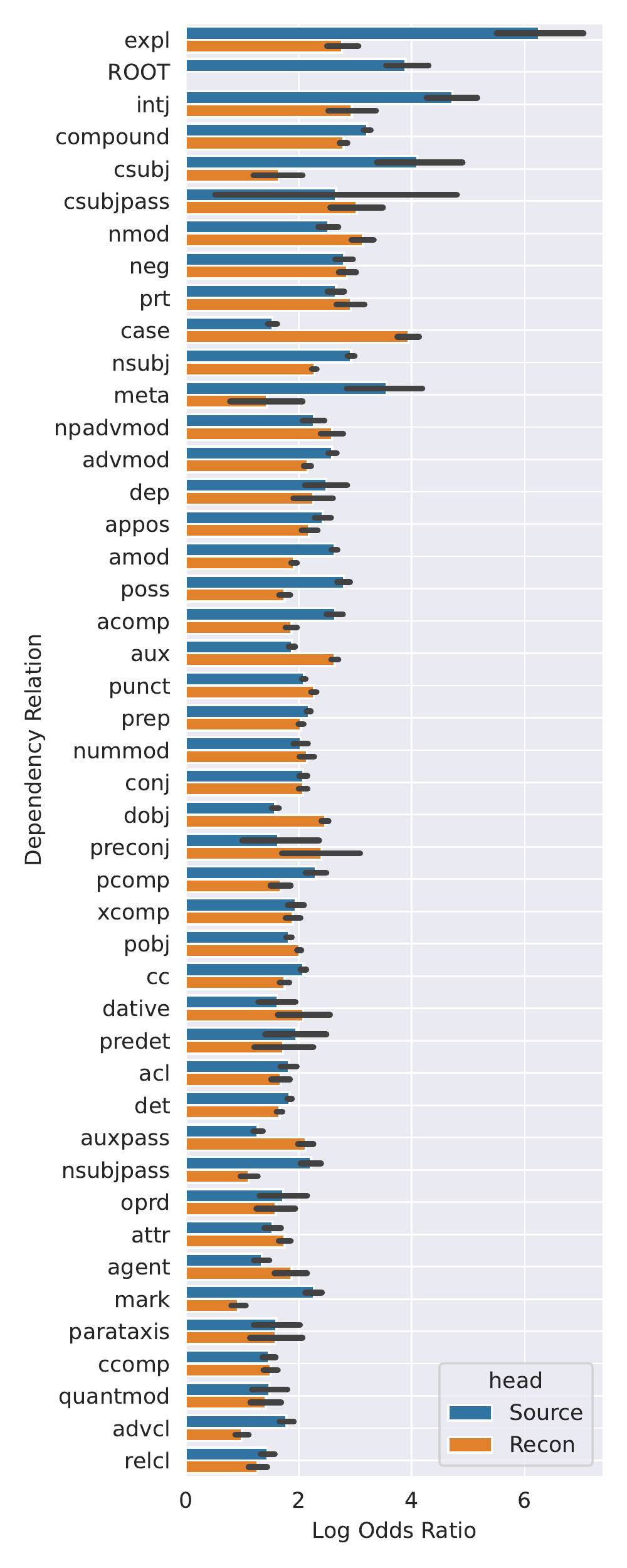}
         \caption{RoBERTa}
    \end{subfigure}%
    \begin{subfigure}[b]{0.3\textwidth}
         \centering
         \includegraphics[width=\textwidth]{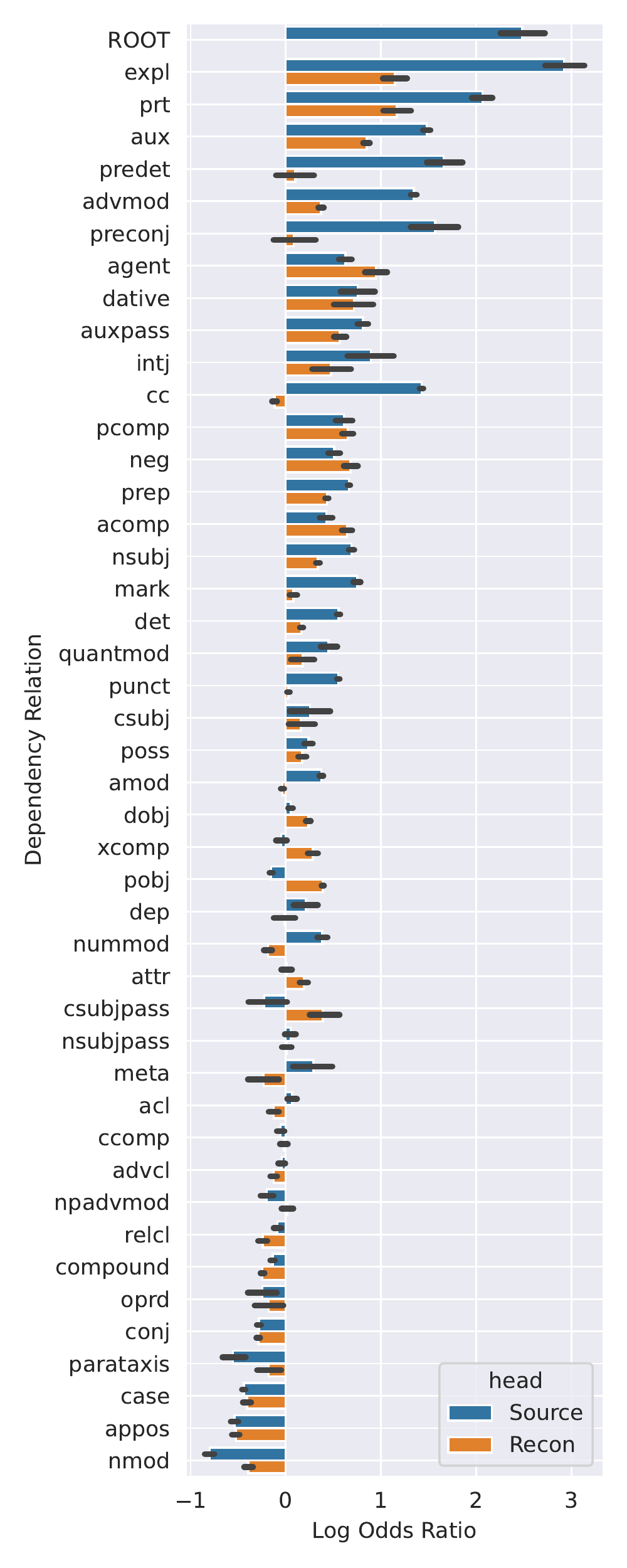}
          \caption{DistilBERT}
    \end{subfigure}

    \caption{Reconstructibility boost (log odds ratio with vs. without source) broken down by the depedency relation label between a \token{source} and \token{recon}.}
    \label{fig:dep-arc}
\end{figure*}

Figure~\ref{fig:dep-arc} shows the full reconstructability boost results for all dependency arc labels in our dataset.

\section{Detailed Decomposition Analysis}
\label{app:decomp-analysis}

\subsection{Creating Ablated Sequences}

\paragraph{Fully contextualized} See Section~\ref{subsec:recon-prob-calculation}.

\paragraph{Static embedding (+position)} We pass through the masked language model the $n$ versions of the input sequence described above, each of which has a single token revealed, at the input layer only. Again, for each \texttt{[MASK]} token in the input sequence, we take the probability of the token in the same position in the original sequence as the reconstruction probability. This value corresponds to the probability of predicting the token in the original sequence given only the static lexical information of the source token and the positional information of the source and recon tokens.

\paragraph{Static embedding (-position)}  We pass through the $n$ single token-revealed versions of the input sequence as described above, but at the input layer, we do not add the positional embeddings. %to create the token embeddings. 
The reconstruction probability obtained, then, corresponds to the probability of predicting the token in the original sequence given only the static lexical information of the source token and no positional information of any of the tokens. 

\paragraph{All mask (+position)} We pass through a fully masked version of the input sequence that consists of the same number of \texttt{[MASK]} tokens and obtain the reconstruction probability of the tokens in the original sequence. Hence, in this scenario, there is no \texttt{source}. The value obtained through this input corresponds to the probability of predicting the token in the original sequence in the absence of any lexical information. Note that the model still has access to the positional embeddings of the \texttt{recon} token, which may still be weakly informative for token prediction.

\paragraph{All mask (-position)} See Section~\ref{subsec:recon-prob-calculation}, `Lexical prior only baseline'.

\begin{table*}[t]
    \centering
    \resizebox{2\columnwidth}{!}{%
    \begin{tabular}{lll}
    \toprule
    \bf  Base & \bf Augmented & \bf What Base vs. Augmented can tell us \\\midrule
    All mask (-position) & All mask (+position) & Reflects the effect of positional information in the absence of any lexical information \\
                                               & & other than the most general lexical priors of the model. \\
    All mask (-position) & Static (-position) & Reflects the effect of static lexical information in the absence of positional information. \\
    All mask (+position) & Static (+position) & Reflects the effect of static lexical information in the presence of positional information. \\
    Static (-position) & Static (+position) & Reflects the effect of positional information in the presence of full lexical information. \\
    Static (+position) & Fully contextualized & Reflects the effect of the contextualization through the layers of the model, beyond the input layer.\\
    All mask (-position) & Fully contextualized & Comparison between the least and most contextualized reconstruction scenarios. Reflects\\
                                               & & the overall change induced by contextualization over the lexical priors of the model. \\
    \bottomrule
    \end{tabular}
    }
    \caption{The comparisons examined and the purpose of the comparisons. Eq.~\ref{eq:lor} is computed by taking the reconstruction probability of a given (token, input sequence) under \textbf{Base} as $q$ and under \textbf{Augmented} as $p$.}
    \label{tab:comparisons}
\end{table*}

\subsection{Representations Compared}

By comparing the reconstruction probabilities described above using Eq.~\ref{eq:lor}, we can gauge the effect of the additional contextual information on performing masked language modeling. For example, if we compare \textbf{Fully contextualized} and \textbf{Static embedding (+position)}, we can quantify the benefit of having the contextualization that happens through applying the model weights to the static representation of the input. If we compare \textbf{Static embedding (+position)} and \textbf{Static embedding (-position)}, we can quantify the benefit of positional embeddings (when given the same static lexical information). We make six different comparisons illustrated in Table~\ref{tab:comparisons}, each comparison serving a different analytic role. 

\subsection{Further Discussion}

\begin{figure*}[h]
    \centering
    \begin{subfigure}[b]{0.3\textwidth}
         \centering
         \includegraphics[width=\textwidth]{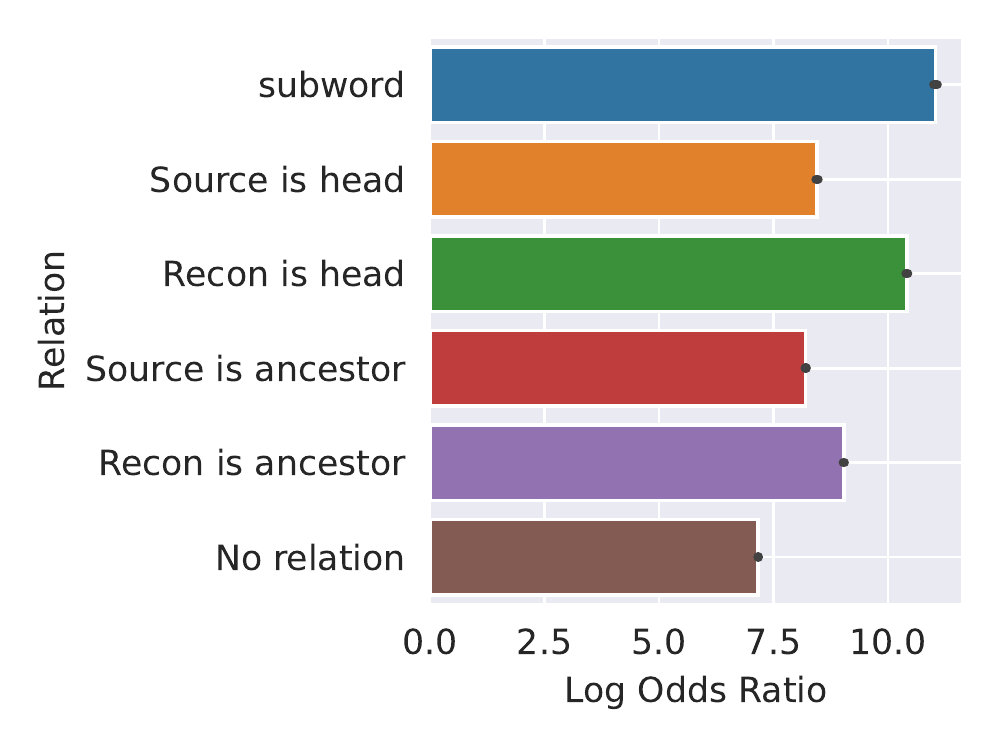}
         \includegraphics[width=\textwidth]{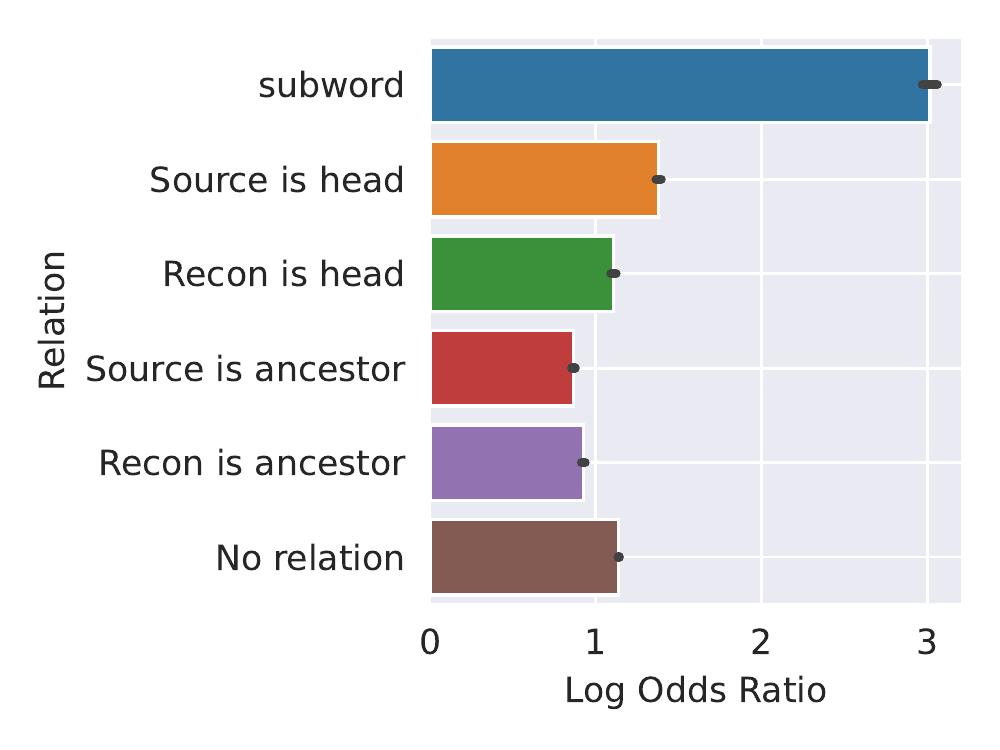}
         \includegraphics[width=\textwidth]{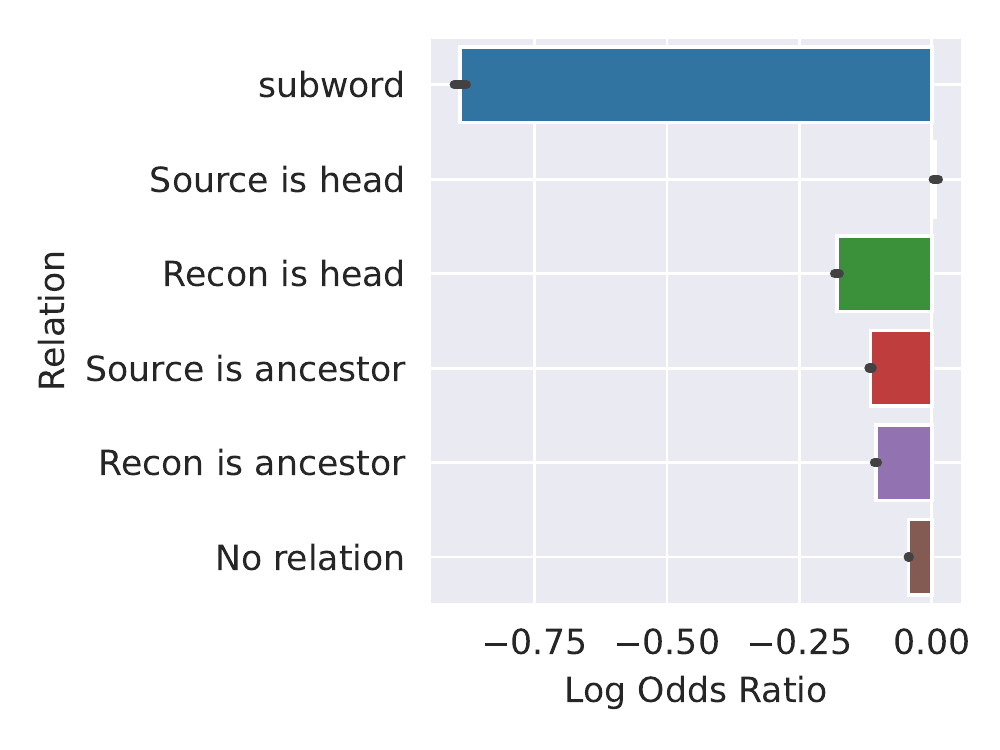}
         \caption{All mask vs. All mask (-position)}
    \end{subfigure}%
    \begin{subfigure}[b]{0.3\textwidth}
         \centering
         \includegraphics[width=\textwidth]{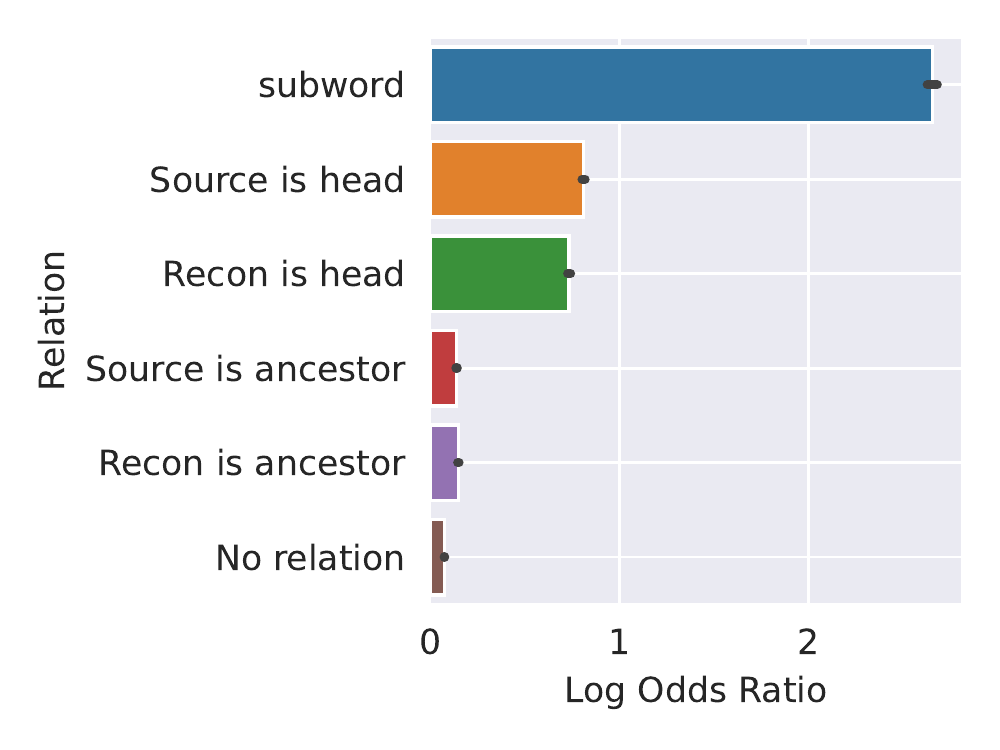}
         \includegraphics[width=\textwidth]{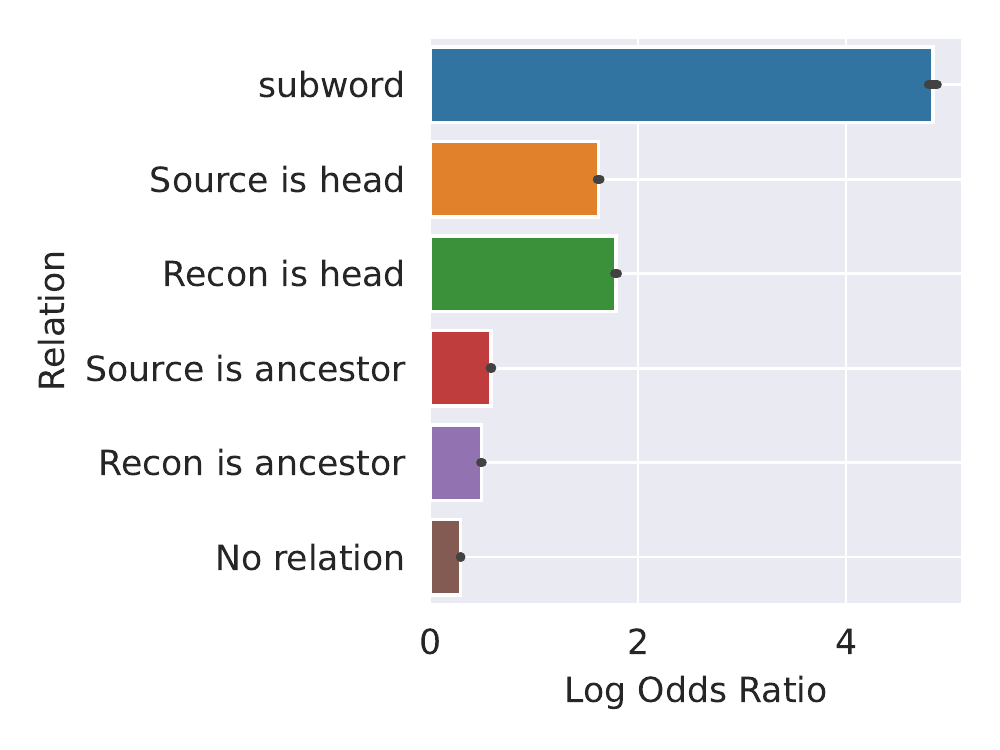}
         \includegraphics[width=\textwidth]{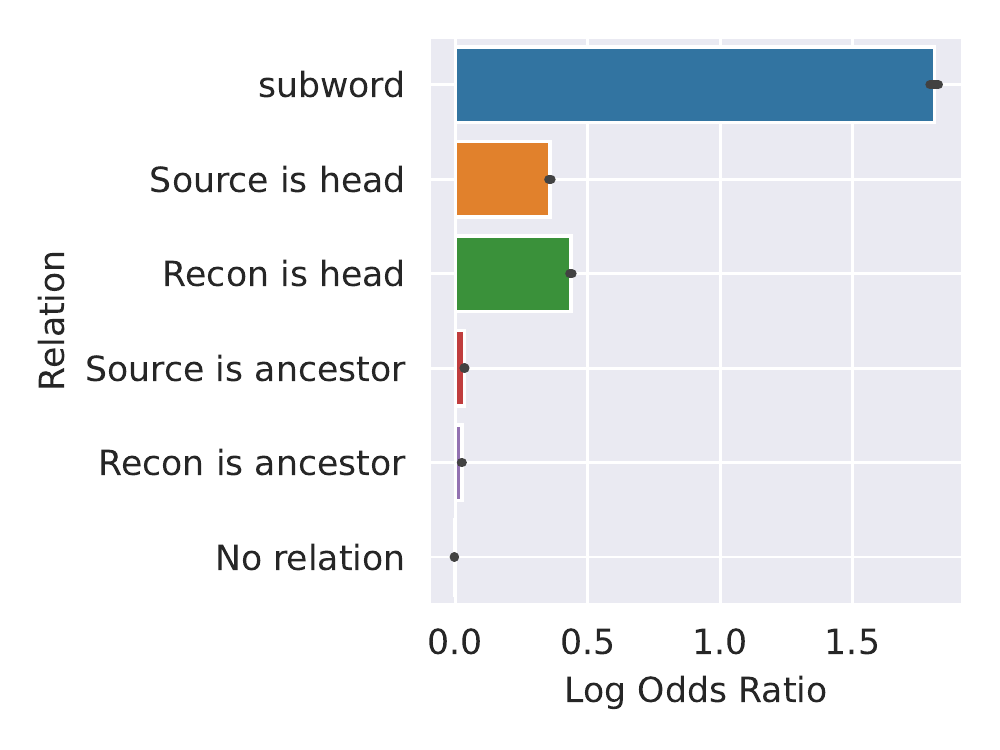}
         \caption{Static vs. All mask}
    \end{subfigure}%
    \begin{subfigure}[b]{0.3\textwidth}
         \centering
         \includegraphics[width=\textwidth]{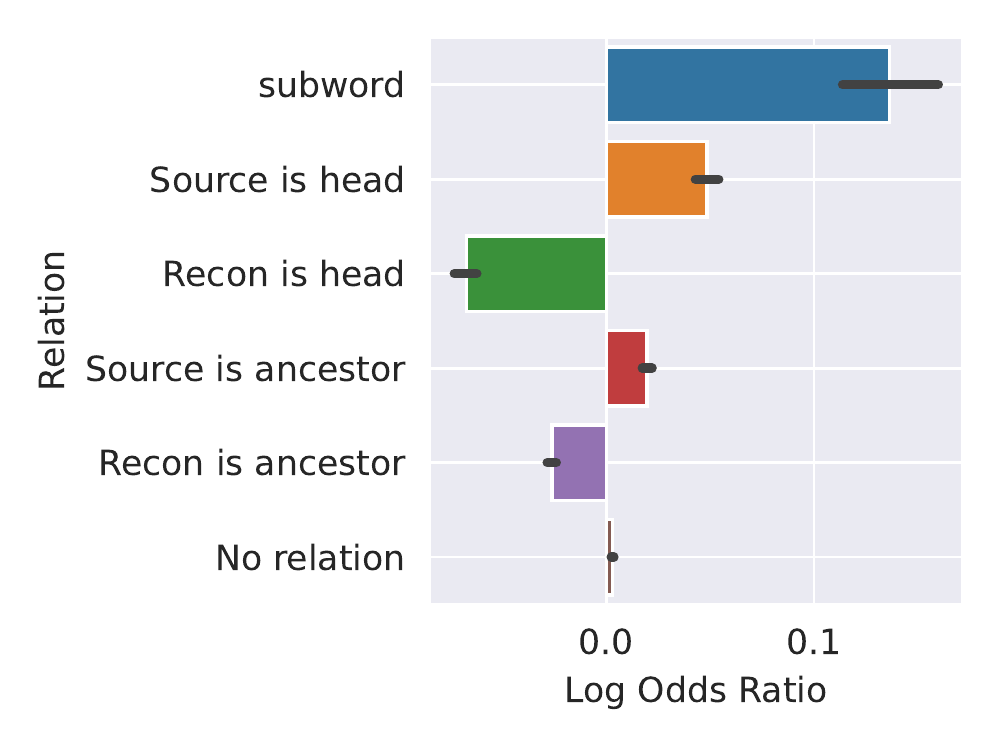}
         \includegraphics[width=\textwidth]{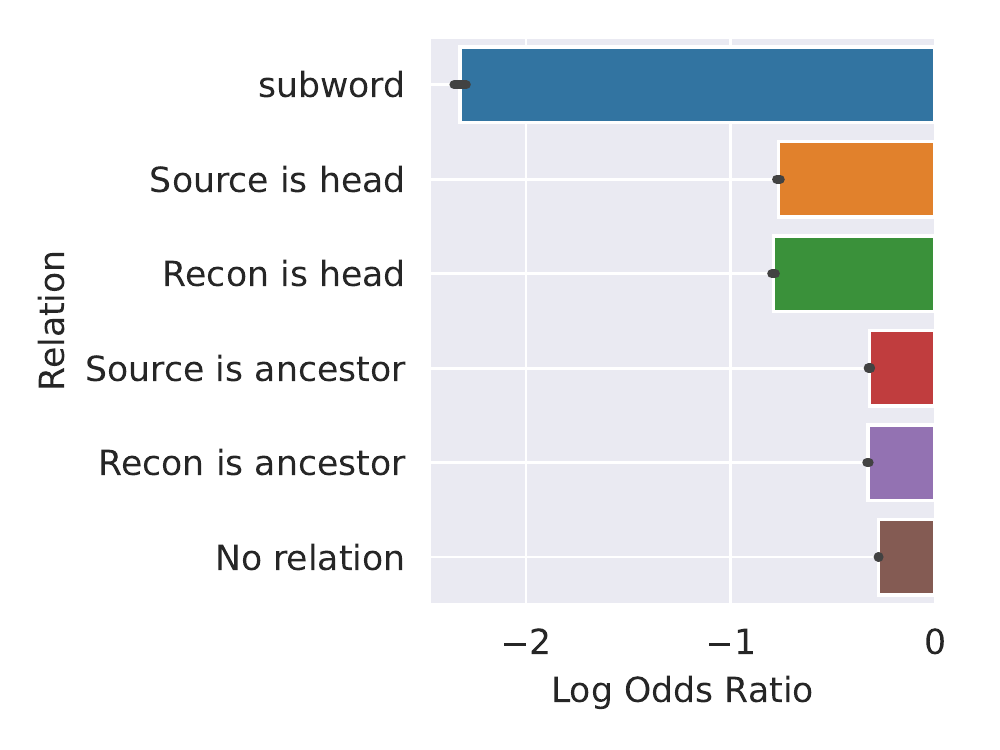}
         \includegraphics[width=\textwidth]{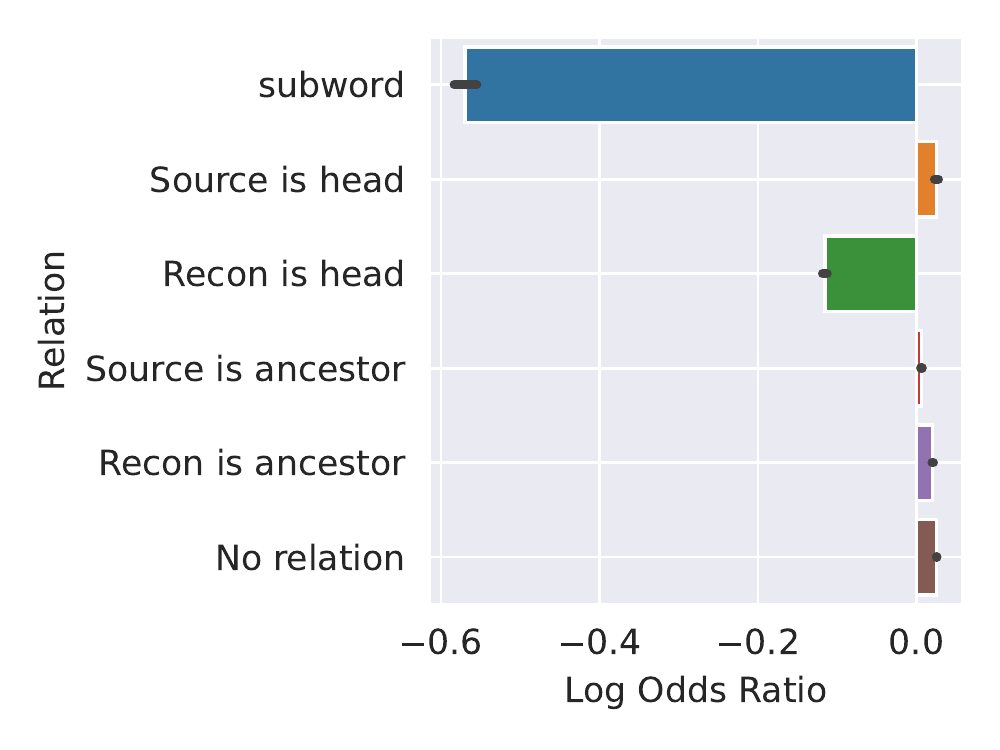}
         \caption{Fully contextualized vs. Static}
    \end{subfigure}
    \caption{Relative reconstructibility (log odds ratio) for BERT (top), RoBERTa (middle), and DistilBERT (bottom).}
    \label{fig:relation-all}
\end{figure*}

We furthermore hypothesized the reconstruction boost from the availability of positional embeddings to be sensitive to the presence of a syntactic relation between \token{source} and \token{recon}. This hypothesis is borne out in BERT and RoBERTA, but not in DistilBERT, suggesting that positional embeddings in DistilBERT are qualitatively different (Figure~\ref{fig:relation-all}, left column). 

\end{document}